%% file: main.tex
\documentclass[letterpaper, 10 pt, journal, twoside]{IEEEtran} 
\pdfoutput=1
\usepackage[T1]{fontenc}
\usepackage[english]{babel}
\usepackage[pdftex]{graphicx}   % for pdf, bitmapped graphics files
\usepackage{epsfig}             % for postscript graphics files
\usepackage{amsmath}            % assumes amsmath package installed
\usepackage{amssymb}            % assumes amsmath package installed
\usepackage{mathtools}
\usepackage{multicol}
\usepackage[bookmarks=true]{hyperref}
\usepackage{xcolor}
\usepackage{siunitx}
\usepackage{lipsum}
\usepackage{scrextend}
\usepackage{algorithm}
\usepackage{algpseudocode}
\usepackage{cite}
\usepackage{multirow}
\usepackage{gensymb}
\usepackage[inkscapelatex=false]{svg}
\usepackage{booktabs}
\usepackage{cleveref} 

\usepackage{bm}

\usepackage{MnSymbol}
\usepackage{mathdots}
\usepackage{amsthm}
\usepackage{empheq}

\input{styles/math}
\input{styles/robotics}
\input{styles/references}

\usepackage{acro}
\usepackage{acro/acro}

\usepackage{siunitx}

\usepackage[caption=false,font=footnotesize]{subfig}

\usepackage{ctable}
\usepackage{makecell}
\usepackage[table]{xcolor}
\definecolor{lightgreen}{RGB}{200,255,200}
\definecolor{lightyellow}{RGB}{255,255,200}
\definecolor{lightred}{RGB}{255,200,200}
\definecolor{lightorange}{RGB}{255,200,150}
\usepackage[dvipsnames]{xcolor}

\title{
Equivariant Filter for Radar-Inertial Odometry
}

\author{% <-this % stops a space
Giulio Delama$^{1}$, Jan Michalczyk$^{1}$, Morten Nissov$^{2}$, Martin Scheiber$^{1}$, Alessandro Fornasier$^{3}$, Kostas Alexis$^{2}$, and Stephan Weiss$^{1}$ \vspace{-8mm}%
\thanks{Manuscript received: September 17, 2025; Revised January 22, 2026 Accepted April 21, 2026.}%
\thanks{This paper was recommended for publication by Editor Pascal Vasseur, upon evaluation of the Associate Editor and Reviewers' comments.
This work was supported by BMK under the grant agreement
FO999925719 (NightWatch) and by the ARO W911NF-21-2-0245. The views and
conclusions contained in this document are those of the authors only. 
The U.S. Government is authorized to reproduce and distribute reprints for Government purposes%
.}%
\thanks{$^{1}$Giulio Delama, Jan Michalczyk, Martin Scheiber, and Stephan Weiss are with the Control of Networked Systems Group, University of Klagenfurt, Austria. {\tt\footnotesize \{name.surname\}@ieee.org}. $^{2}$Morten Nissov and Konstantinos Alexis are with the Autonomous Robots Lab, Norwegian University of Science and Technology, Norway. {\tt\footnotesize \{name.surname\}@ntnu.no}. $^{3}$Alessandro Fornasier is with Hexagon Robotics, Zurich, Switzerland. {\tt\footnotesize alessandrofornasierphd@gmail.com}}
\thanks{Digital Object Identifier (DOI): see top of this page.}
}

\begin{document}
\bstctlcite{BSTcontrol}
\markboth{IEEE Robotics and Automation Letters. Preprint Version. Accepted April, 2026}
{Delama \MakeLowercase{\textit{et al.}}: Equivariant Filter for Radar-Inertial Odometry} 
\maketitle
%\thispagestyle{empty}
%\pagestyle{empty}

%%%%%%%%%%%%%%%%%%%%%%%%%%%%%%%%%%%%%%%%%%%%%%%%%%%%%%%%%%%%%%%%%%%%%%%%%%%%%%%%

\input{sections/abstract}
\input{sections/introrelated}
\input{sections/math}
\input{sections/filter}
\input{sections/multistate}
\input{sections/results}
\input{sections/conclusion}

\bibliographystyle{IEEEtran}
\bibliography{bibliography/EqF.bib, bibliography/preintegration.bib, bibliography/radar.bib, bibliography/extra.bib}
%%%%%%%%%%%%%%%%%%%%%%%%%%%%%%%%%%%%%%%%%%%%%%%%%%%%%%%%%%%%%%%%%%%%%%%%%%%%%%%%

\end{document}

%% file: styles/math.tex
\theoremstyle{plain}
\newtheorem{theorem}{Theorem}[section]

\newtheorem{lemma}[theorem]{Lemma}

\theoremstyle{definition}

\theoremstyle{remark}

\newcommand{\R}{\mathbb{R}}
\newcommand{\norm}[1]{\left\lVert #1 \right\rVert}

\newcommand{\inv}[1]{{#1}^{-1}}

\newcommand{\grpG}{\mathbf{G}}
\newcommand{\gothg}{\mathfrak{g}}

\newcommand{\twoel}[2]{\left(#1,\; #2\right)}
\newcommand{\threeel}[3]{\left(#1,\; #2,\; #3\right)}
\newcommand{\fourel}[4]{\left(#1,\; #2,\; #3,\; #4\right)}

\newcommand{\Adsym}[2]{\mathrm{Ad}_{#1}\!\left[#2\right]}

\newcommand{\AdMsym}[1]{\mathbf{Ad}^\vee_{#1}}

\newcommand{\SE}[1]{\mathbf{SE}(#1)}
\newcommand{\se}[1]{\mathfrak{se}(#1)}

\newcommand{\SEtwo}[1]{\mathbf{SE}_2(#1)}
\newcommand{\setwo}[1]{\mathfrak{se}_2(#1)}

\newcommand{\G}[1]{\mathbf{Gal}(#1)}
\newcommand{\g}[1]{\mathfrak{gal}(#1)}

\newcommand{\TG}[2]{{#1}_{#2}^\ltimes}

\newcommand{\Jl}[1]{\mathbf{J}_{\mathrm{L}}(#1)}

\newcommand{\Diff}[3]{\mathrm{D}_{#2}|_{#3}{#1}(#2)}
\newcommand{\diff}[1]{\mathrm{d}{#1}}
\newcommand{\difff}[2]{\mathrm{d}{#1}(#2)}

\newcommand{\PI}[2]{\Pi_{\scalebox{0.5}{$#1$}}(#2)}
\newcommand{\PIg}[1]{\Pi_{\scalebox{0.5}{$#1$}}}
\DeclareMathOperator{\atantwo}{atan2}

%% file: styles/robotics.tex
\newcommand{\Pose}[2]{\prescript{#1}{}{\mathbf{T}}_{#2}}
\newcommand{\Rot}[2]{\prescript{#1}{}{\mathbf{R}}_{#2}}

\newcommand{\Vector}[3]{\prescript{#1}{}{\bm{#2}}_{#3}}

\newcommand{\dotPose}[2]{\dot{\prescript{#1}{}{\mathbf{T}}_{#2}}}
\newcommand{\dotRot}[2]{\dot{\prescript{#1}{}{\mathbf{R}}_{#2}}}
\newcommand{\dotVector}[3]{\prescript{#1}{}{\dot{\bm{#2}}}_{#3}}

\newcommand{\hatVector}[3]{\prescript{#1}{}{\hat{\bm{#2}}}_{#3}}

\newcommand{\curlVector}[3]{\prescript{#1}{}{\tilde{\bm{#2}}}_{#3}}

\newcommand{\ringVector}[3]{\prescript{#1}{}{\mathring{\bm{#2}}}_{#3}}

\newcommand{\vel}[2]{\Vector{#1}{v}{#2}}
\newcommand{\pos}[2]{\Vector{#1}{p}{#2}}
\newcommand{\bias}[2]{\Vector{#1}{b}{#2}}

\newcommand{\dotvel}[2]{\dotVector{#1}{v}{#2}}
\newcommand{\dotpos}[2]{\dotVector{#1}{p}{#2}}
\newcommand{\dotbias}[2]{\dotVector{#1}{b}{#2}}

\newcommand{\wInp}[2]{\Vector{#1}{w}{#2}}
\newcommand{\tauInp}[2]{\Vector{#1}{\tau}{#2}}
\newcommand{\omegaInp}[2]{\Vector{#1}{\omega}{#2}}
\newcommand{\aInp}[2]{\Vector{#1}{a}{#2}}
\newcommand{\nuInp}[2]{\Vector{#1}{\nu}{#2}}

\newcommand{\curlomegaInp}[2]{\curlVector{#1}{\omega}{#2}}
\newcommand{\curlaInp}[2]{\curlVector{#1}{a}{#2}}
\newcommand{\curlwInp}[2]{\curlVector{#1}{w}{#2}}

\newcommand{\ringtauInp}[2]{\ringVector{#1}{\tau}{#2}}

\newcommand{\delt}{\delta t}

\newcommand{\eyen}[1]{\mathbf{I}_{#1}}

\newcommand{\zeronm}[2]{\mathbf{0}_{#1 \times #2}}

\newcommand{\Rn}[1]{\R^{#1}}

\newcommand{\Rnm}[2]{\R^{#1 \times #2}}

\newcommand{\point}{\;\cdot\;}

\newcommand{\vd}{v_{\scalebox{0.5}{\textnormal{D}}}}

\newcommand{\pf}{\Vector{}{p}{\scalebox{0.5}{$f$}}}

\newcommand{\pfi}{\Vector{}{p}{\scalebox{0.5}{$f$}i}}

\newcommand{\h}[1]{h^{\scalebox{0.6}{$#1$}}}

\newcommand{\Lift}[1]{\Lambda_{\scalebox{0.5}{$#1$}}}

\newcommand{\stateaction}[1]{\phi_{\scalebox{0.5}{$#1$}}}

\newcommand{\gn}{\Vector{}{g}{\scalebox{0.5}{\textnormal{N}}}}

\newcommand{\wn}{\Vector{}{w}{\scalebox{0.5}{\textnormal{N}}}}

\newcommand{\curlwn}{\curlVector{}{w}{\scalebox{0.5}{\textnormal{N}}}}

\newcommand{\ringwn}{\ringVector{}{w}{\scalebox{0.5}{\textnormal{N}}}}

\newcommand{\Matrix}[3]{\mathbf{#1}^{\scalebox{0.6}{$#2$}}_{#3}}

%% file: styles/references.tex
\newcommand{\figref}[1]{Fig.~\ref{fig:#1}}
\newcommand{\secref}[1]{Sec.~\ref{sec:#1}}

\newcommand{\tabref}[1]{Tab.~\ref{tab:#1}}
\newcommand{\algoref}[1]{Alg.~\ref{alg:#1}}

%% file: sections/abstract.tex
\begin{abstract}
Radar-Inertial Odometry (RIO) based on the Extended Kalman Filter (EKF) relies on accurate extrinsic calibration between the radar and the Inertial Measurement Unit (IMU) and is sensitive to disturbances, as large linearization errors can degrade performance or even cause divergence.
To address these limitations, this letter proposes an Equivariant Filter (EqF) for RIO based on a Lie group symmetry that geometrically couples navigation states and IMU biases, extending it to incorporate radar-IMU extrinsic calibration and multi-state constraint updates.
This equivariant formulation inherently preserves consistency and enhances robustness, enabling reliable state estimation even under poor or completely wrong initialization of calibration states.
Real-world experiments on two different Uncrewed Aerial Vehicles (UAVs) show that the proposed EqF-RIO achieves state-of-the-art accuracy under correct extrinsic calibration and offers improved convergence under large calibration errors, where the conventional EKF-RIO fails.
Evaluation code is open-sourced.
\end{abstract}

\begin{IEEEkeywords}
Localization, Sensor Fusion, Range Sensing
\end{IEEEkeywords}
\vspace{-4mm}

\IEEEpeerreviewmaketitle

%% file: sections/introrelated.tex
\section{Introduction and Related Work}
\IEEEPARstart{A}{ccurate} and reliable localization in Global Navigation Satellite System (GNSS)-denied environments remains a critical challenge for autonomous navigation.
Radar-aided Inertial Navigation Systems (INSs) have emerged as a promising solution, leveraging complementary information from millimeter-wave radars and Inertial Measurement Units (IMUs) to estimate the motion of a mobile robot.
This method is commonly known as Radar-Inertial Odometry (RIO) and remains largely unaffected by adverse environmental conditions, unlike camera-based or Light Detection and Ranging (LiDAR)-based approaches.
Its resilience to challenging factors, such as poor illumination, fog, rain, snow, and airborne particulates like dust or smoke, arises from the inherent properties of radio waves, highlighting its potential for accurate and reliable localization in difficult scenarios~\cite{jan_ssrr, Nissov2024DegradationOdometry}.

Among recent approaches to RIO, methods based on the nonlinear optimization of factor graphs~\cite{Nissov2024DegradationOdometry, Michalczyk2024Tightly-CoupledOdometry, lessismore, Xu2025IncorporatingSLAM, cynoh-2025-icra} are generally considered more reliable than filter-based approaches, owing to iterative re-linearization of the measurement and motion models around the current state estimate.
Authors in~\cite{Nissov2024DegradationOdometry} introduce a factor graph estimator for multi-modal fusion, combining LiDAR feature factors, radar velocity factors computed via RANSAC least squares, and IMU preintegration factors.
They demonstrate the enhanced robustness that radar provides in typical LiDAR-degenerate environments, such as fog-filled corridors and geometrically self-similar tunnels, for both RIO and LiDAR-RIO approaches.
In~\cite{Michalczyk2024Tightly-CoupledOdometry}, the authors design radar velocity factors in a tightly-coupled manner, where each factor corresponds to a single Doppler velocity measurement.
Additional factors based on 3D radar points and persistent features are also included.
The method presented in~\cite{lessismore} employs radar velocity and 3D point factors, and leverages object reflectivity properties in feature tracking.
This approach is further refined in~\cite{Xu2025IncorporatingSLAM} by representing 3D point uncertainty in spherical coordinates, thereby enabling more accurate modeling of radar measurement noise.
The latest study~\cite{cynoh-2025-icra} leverages gravity estimation to mitigate vertical drift.

\input{figures/first}
Despite advantages in terms of reliability and robustness, the high computational cost of optimization-based approaches can limit their real-time deployment on resource-constrained platforms such as Uncrewed Aerial Vehicles (UAVs).
As a more resource-efficient solution, Extended Kalman Filter (EKF)-based methods for RIO~\cite{doer2021radar, jan_ssrr, Michalczyk2022Radar-InertialManoeuvres, Michalczyk2022Tightly-CoupledOdometry, Michalczyk2023Multi-StateLandmarks, kim2025} provide a practical alternative to optimization-based approaches.
In a recent work~\cite{doer2021radar}, IMU data is fused with radar Doppler velocity measurements within an EKF framework.
Later work~\cite{Michalczyk2022Radar-InertialManoeuvres, jan_ssrr} introduces an EKF-RIO using radar distance measurements to fixed landmarks in the update step.
The same authors further develop a tightly-coupled EKF-RIO in~\cite{Michalczyk2022Tightly-CoupledOdometry}, fusing IMU data with radar Doppler velocity and 3D point measurements.
In particular, their method employs stochastic cloning~\cite{Roumeliotis2002StochasticMeasurements} to enable association of tracked 3D radar points across consecutive scans, following an approach similar to a Multi-State Constraint Kalman Filter~\cite{Mourikis2007ANavigation}.
This framework is further extended in~\cite{Michalczyk2023Multi-StateLandmarks} to include radar persistent features.
Latest research~\cite{kim2025} introduces the online estimation of the temporal offset between IMU and radar measurements.
Filter-based approaches provide accurate and computationally efficient state estimation, making them attractive for real-time applications.
However, they are sensitive to calibration errors and unmodeled effects, which can introduce large linearization errors from which the EKF may not recover, resulting in degraded performance or failure.

Recently, Equivariant Filters (EqFs)~\cite{VanGoor2020EquivariantSpaces, vanGoor2022EquivariantEqF, Fornasier2022EquivariantBiases, Fornasier2022OvercomingCalibration, Scheiber2023RevisitingApproach, Fornasier2023MSCEqF:Navigation, Fornasier2024AnSystem} have emerged as a leading method in modern state estimation, drawing increasing attention for their improved robustness, faster convergence, and larger basin of attraction compared to conventional EKFs.
By exploiting \emph{equivariant symmetries} of the system dynamics based on Lie groups~\cite{Fornasier2024EquivariantNavigation, Fornasier2025EquivariantSystems, Delama2025EquivariantApproach}, EqFs naturally preserve the geometric structure of the state space, thus reducing linearization errors and maintaining consistency.
As a result, they are less sensitive to sensor calibration inaccuracies and can recover from external perturbations that would typically cause standard EKFs to diverge.
In this sense, EqFs inherit much of the robustness commonly associated with optimization-based methods, while fully retaining the computational efficiency of filtering.
This unique balance of efficiency and reliability makes EqFs particularly well-suited for RIO.

In this letter, we derive a novel \emph{discrete-time EqF} for RIO with radar Doppler velocities and point-cloud measurements.
Drawing from the state-of-the-art biased-INS formulation that leverages the \emph{Galilean} group~\cite{Delama2025EquivariantApproach}, we extend this symmetry to incorporate extrinsic calibration states (\secref{ins}, \ref{sec:symmetry}).
Analytical discrete-time linearization matrices ensure computational efficiency by avoiding numerical differentiation or matrix exponentials typically required in continuous-time discretization (\secref{error}).
We enhance filter consistency by modeling 3D point uncertainty in spherical coordinates, which more accurately captures radar point-cloud noise characteristics~\cite{Xu2025IncorporatingSLAM}, and by including angular velocity noise in the Doppler velocity measurement model (\secref{doppler_update}).
Finally, we extend our filter with \emph{multi-state constraint} updates (\secref{augment}, \ref{sec:msc_update}).

The proposed EqF-RIO is validated on real-world datasets collected from two different UAVs, shown in~\figref{UAVs}.
Our method is compared against a state-of-the-art tightly-coupled EKF-RIO~\cite{Michalczyk2022Tightly-CoupledOdometry}, as both filters share the same updates but differ fundamentally in their formulation.
Experimental results show that EqF-RIO achieves state-of-the-art accuracy under precise radar-IMU extrinsic calibration while preserving stable convergence under large initial calibration errors, outperforming~\cite{Michalczyk2022Tightly-CoupledOdometry} in robustness.
The evaluation code is made publicly available.

In summary, the main contributions of this letter are:
\begin{itemize}
\item A novel discrete-time, multi-state constraint EqF for radar-aided INSs with radar-IMU extrinsic calibration, offering improved robustness and faster convergence compared to conventional EKF-based approaches for RIO.
\item Analytical discrete-time filter linearization matrices derived from a state-of-the-art Galilean group symmetry for the biased-INS, enabling efficient filter implementation.
\item Extensive real-world validation on two datasets collected with different UAVs, demonstrating superior performance over a state-of-the-art EKF-RIO~\cite{Michalczyk2022Tightly-CoupledOdometry}, particularly under large rotational calibration perturbations.
\end{itemize}

%% file: figures/first.tex
\begin{figure}
    \centering
    \includegraphics[width=\linewidth]{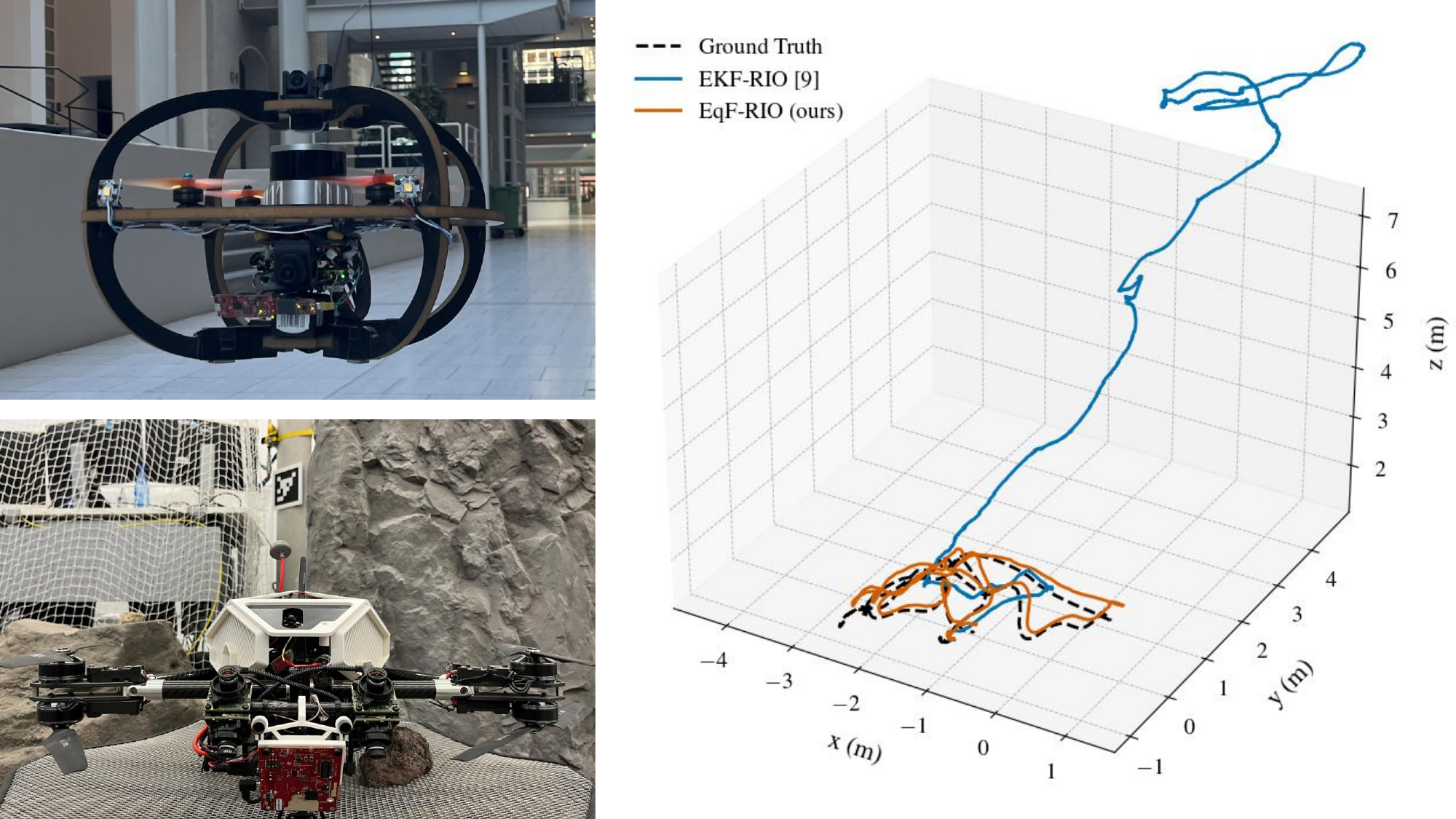}
    \vspace{-7mm}
    \caption{The two UAV platforms used to collect the real-world flight datasets for evaluating the proposed EqF-RIO, along with a comparison against a state-of-the-art EKF-RIO~\cite{Michalczyk2022Tightly-CoupledOdometry}. On the top left is a self-built UAV, and on the bottom left is the ARDEA-X~\cite{ardea}. The right plot shows results on the \texttt{ardea-2} sequence from the ARDEA-X dataset. Our EqF-RIO rapidly converges and recovers from an initial $80^\circ$ radar–IMU rotation calibration error around the y-axis, whereas EKF-RIO fails to recover under the same initial conditions.}% This highlights the robustness and great convergence properties of our method.}
    \label{fig:UAVs}
    \vspace{-5mm}
\end{figure}

%% file: sections/math.tex
\section{Notation and Mathematical Preliminaries}
This letter utilizes bold lowercase letters to represent vectors, bold capital letters to denote matrices, and regular letters to indicate elements within a symmetry group.
The $n$-dimensional identity matrix is denoted ${\eyen{n} \in \Rnm{n}{n}}$, and the ${n \times m}$ zero matrix is denoted ${\zeronm{n}{m} \in \Rnm{n}{m}}$.
Estimates are denoted $\hat{x}$, and noisy measurements are denoted $\tilde{x}$.
The \emph{Fréchet} derivative (\emph{differential}) of a differentiable function ${F(\xi) \in \mathcal{M}}$ evaluated at ${\xi = \bar{\xi}}$ in the direction ${\eta \in \mathrm{T}_{F(\bar{\xi})}\mathcal{M}}$ is denoted ${\difff{F}{\bar{\xi}}[\eta] = \Diff{F}{\xi}{\bar{\xi}}[\eta]}$.
In the literature, the simplified notation ${\diff{F}[\eta] \coloneqq \difff{F}{I}[\eta] = \Diff{F}{\xi}{I}[\eta]}$ is often used for differentials evaluated at the identity $I$ or at other trivial or implicit points.
To avoid confusion, we adopt the clearer and more explicit notation ${\Diff{F}{\xi}{\bar{\xi}}[\eta]}$ for general differential, and the more concise notation ${\diff{F}[\eta]}$ only for those evaluated specifically at identity.
For brevity, a function $F(x,y,...)$ may be written as $F(x)$ when other arguments are fixed or implicit.\vspace{-1mm}

\subsection{Left-trivialized tangent group}
\label{sec:TangentGroup}
Given a Lie group $\grpG$ and its Lie algebra $\gothg$, the set of pairs ${\twoel{X}{u} \in \grpG \times \gothg}$ corresponding to ${\twoel{X}{u} \mapsto \mathrm{dL}_X[u] \in \mathrm{T}_X \grpG}$ can be given a group structure called the \emph{left-trivialized tangent group} of $\grpG$ and denoted $\TG{\grpG}{\gothg} \coloneqq \grpG \ltimes \gothg$.
Let ${X, Y \in \grpG \ltimes \gothg}$ where ${X = \twoel{A}{a}}$ and ${Y = \twoel{B}{b}}$, ${A, B \in \grpG}$ and ${a, b \in \gothg}$.
The group operation, i.e., the \emph{semi-direct product}, and the inverse are
\begin{align}
&\!\!\!\!XY = \twoel{AB}{a + \Adsym{A}{b}}, &&\inv{X} = \twoel{\inv{A}}{-\Adsym{\inv{A}}{a}},
\end{align}
and the identity element is $\twoel{I}{0}$.
For a brief introduction to the \emph{Galilean} group $\G{3}$, readers can refer to~\cite{Delama2025EquivariantApproach}, while a comprehensive overview of \emph{equivariant symmetries} in the context of aided inertial navigation can be found in~\cite{Fornasier2024EquivariantNavigation, Fornasier2025EquivariantSystems}.\vspace{-1mm}

\subsection{Projection operator and useful maps}
\label{sec:maps}
Consider two matrix Lie groups $\grpG_1, \grpG_2$ with Lie algebras $\gothg_1, \gothg_2$.
We define the \emph{projection operators} $\PIg{\grpG_1}$ and $\PIg{\gothg_1}$ as
\begin{subequations}
    \begin{align}
        \PIg{\grpG_1} &: \grpG_2 \to \grpG_1, &&\PI{\grpG_1}{X} = Y \in \grpG_1 &&&\forall X \in \grpG_2 , \\
        \PIg{\gothg_1} &: \gothg_2 \to \gothg_1, &&\PI{\gothg_1}{u} = v \in \gothg_1 &&&\forall u \in \gothg_2 . 
    \end{align}
\end{subequations}
If ${\grpG_1 \subset \grpG_2}$, the projection ${\PI{\grpG_1}{X}}$ extracts from ${X \in \grpG_2}$ the sub-elements matching the structure of $\grpG_1$.
Similarly, if ${\gothg_1 \subset \gothg_2}$, the projection ${\PI{\gothg_1}{u}}$ extracts from ${u \in \gothg_2}$ the part corresponding to $\gothg_1$.
If ${\grpG_1 \supset \grpG_2}$, the projection ${\PI{\grpG_1}{X}}$ maps ${X \in \grpG_2}$ into $\grpG_1$ by completing it with group identity elements in the complementary structure.
Similarly, if ${\gothg_1 \supset \gothg_2}$, the projection ${\PI{\gothg_1}{u}}$ maps ${u \in \gothg_2}$ into $\gothg_1$ by appending zeros.

Let ${X\!=\!\fourel{A}{a}{b}{c}\!\in\!\G{3}}$, ${u\!=\!\fourel{\Vector{}{\omega}{}^\wedge}{\Vector{}{\alpha}{}}{\Vector{}{\beta}{}}{\gamma}\!\in\!\g{3}}$, ${Y\!=\!\threeel{A}{a}{b} \in \SEtwo{3}}$, and ${v\!=\!\threeel{\Vector{}{\omega}{}^\wedge}{\Vector{}{\alpha}{}}{\Vector{}{\beta}{}} \in \setwo{3}}$.
Using the \emph{projection operators}, we define the following linear maps:%
\begin{subequations}
    \begin{align}
        \PI{\SEtwo{3}}{X} &= \threeel{A}{a}{b} \in \SEtwo{3} , \\
        \PI{\SE{3}}{X} &= \PI{\SE{3}}{Y} = \twoel{A}{b} \in \SE{3} , \label{eq:PISE3}\\
        \PI{\G{3}}{Y} &= \fourel{A}{a}{b}{0} \in \G{3} , \\
        \PI{\g{3}}{v} &= \fourel{\Vector{}{\omega}{}^\wedge}{\Vector{}{\alpha}{}}{\Vector{}{\beta}{}}{0} \in \g{3} .
    \end{align}
\end{subequations}
Note that these maps can be applied in the same way to elements of the G-torsor, i.e., the underlying \emph{manifold}.
For example, considering an extended pose ${\mathbf{T} = \threeel{\mathbf{R}}{\Vector{}{v}{}}{\Vector{}{p}{}} \in \mathcal{SE}_2(3)}$, one can extract the pose as $\PI{\mathcal{SE}(3)}{\mathbf{T}} = \twoel{\mathbf{R}}{\Vector{}{p}{}} \in \mathcal{SE}(3)$.

For all ${\mathbf{M} \in \Rnm{9}{N}}$ we define the map ${\Xi : \Rnm{9}{N} \to \Rnm{6}{N}}$ as
\begin{align}
    &\!\!\!\!\Xi(\mathbf{M}) = \Xi\left(\begin{bmatrix} \mathbf{M}^{(1)} \\ \mathbf{M}^{(2)} \\ \mathbf{M}^{(3)} \end{bmatrix}\right) = \begin{bmatrix} \mathbf{M}^{(1)} \\ \mathbf{M}^{(3)} \end{bmatrix} \in \Rnm{6}{N} , &&\mathbf{M}^{(i)} \in \Rnm{3}{N} .
    \label{eq:Xi}
\end{align}
For all ${\mathbf{M} \in \Rnm{10}{10}}$ we define the map ${\Theta : \Rnm{10}{10} \to \Rnm{9}{9}}$ as
\begin{equation}
    \Theta(\mathbf{M}) = \Theta\left(\begin{bmatrix} \mathbf{M}^{(1)} & ... \\ ... & ... \end{bmatrix}\right) = \mathbf{M}^{(1)} \in \Rnm{9}{9} .
    \label{eq:Theta}
\end{equation}
For all ${\mathbf{M} \in \Rnm{10}{10}}$ we define the map ${\Omega : \Rnm{10}{10} \to \Rnm{9}{10}}$ as
\begin{equation}
    \Omega(\mathbf{M}) = \Omega\left(\begin{bmatrix} \mathbf{M}^{(1)} \\ ... \end{bmatrix}\right) = \mathbf{M}^{(1)} \in \Rnm{9}{10} .
    \label{eq:Omega}
\end{equation}

%% file: sections/filter.tex
\section{Discrete-Time Equivariant Filter Design}
\label{sec:filter}
The main contribution of this letter is a novel \emph{discrete-time} EqF for tightly coupled fusion of inertial data with radar point-cloud and Doppler velocity measurements.
We provide a complete step-by-step derivation of the filter with all components in closed form, along with a practical implementation detailed in~\algoref{eqf}.
This section introduces the core EqF formulation for RIO with Doppler velocity updates only, while~\secref{msc} extends it by including \emph{multi-state constraint} updates.

\subsection{Radar-aided Inertial Navigation System (radar-INS)}
\label{sec:ins}
Consider a mobile robot equipped with an IMU that delivers biased angular velocity and linear acceleration measurements and a radar sensor providing 3D point-cloud and Doppler velocity measurements.
The continuous-time and noise-free radar-INS is defined by the following equations:
\vspace{-2mm}\\
\begin{subequations}
\centering
\hspace{-5mm}
\begin{minipage}{0.25\textwidth}
    \begin{align}
      \dotRot{}{} &= \Rot{}{}\left(\omegaInp{}{} - \bias{}{\omega}\right)^\wedge,\\
      \dotvel{}{} &= \Rot{}{}\left(\aInp{}{} - \bias{}{a}\right) + \Vector{}{g}{},\\
      \dotpos{}{} &= \vel{}{},
    \end{align}
\end{minipage}
\hspace{10mm}
\begin{minipage}{0.15\textwidth}
    \begin{align}
      \dotbias{}{\omega} &= \tauInp{}{\omega},\\
      \dotbias{}{a} &= \tauInp{}{a},\\
      \dot{\mathbf{S}} &= \mathbf{S}\Vector{}{\mu}{S}^\wedge,\\
      \dot{\Vector{}{t}{}} &= \Vector{}{\mu}{t},
    \end{align}
\end{minipage}
\label{eq:ins}
\vspace{2mm}
\end{subequations}\\
where ${\Rot{}{} \in \mathcal{SO}(3)}$ and ${\vel{}{}, \pos{}{} \in \Rn{3}}$ denote the \emph{core states} (or \emph{navigation states}), i.e., the rigid body orientation, velocity, and position in the global reference frame.
${\bias{}{\omega}, \bias{}{a} \in \Rn{3}}$ denote the IMU \emph{bias states}, ${\Vector{}{g}{} = \threeel{0}{0}{-9.81} \in \Rn{3}}$ is the gravity vector, and ${\omegaInp{}{}, \aInp{}{} \in \Rn{3}}$ are the \emph{biased} rigid body angular velocity and linear acceleration.
The \emph{calibration states} include the rotation ${\mathbf{S} \in \mathcal{SO}(3)}$ and the translation ${\Vector{}{t}{} \in \Rn{3}}$ of the radar sensor in the IMU reference frame.
${\tauInp{}{\omega}, \tauInp{}{a}, \Vector{}{\mu}{S}, \Vector{}{\mu}{t} \in \Rn{3}}$ are inputs used to model the evolution of the bias terms and calibration, e.g., they are zero if those variables are modeled as constant quantities.
By extending~\eqref{eq:ins} with additional virtual inputs and bias states~\cite{Fornasier2025EquivariantSystems}, the radar-INS can be reformulated as follows.

Let ${\xi = \threeel{\Pose{}{}}{\bias{}{}}{\mathbf{L}} \in \mathcal{M} \coloneqq \mathcal{SE}_2(3) \times \Rn{9} \times \mathcal{SE}(3)}$ define the state of the augmented system.
The core states are modeled as extended poses ${\Pose{}{} = \threeel{\Rot{}{}}{\vel{}{}}{\pos{}{}} \in \mathcal{SE}_2(3)}$.
The bias states ${\bias{}{} = \threeel{\bias{}{\omega}}{\bias{}{a}}{\bias{}{\nu}} \in \Rn{9}}$ include the IMU biases and an additional \emph{virtual} velocity bias ${\bias{}{\nu} \in \Rn{3}}$ introduced in~\cite{Fornasier2022EquivariantBiases}.
The calibration states are denoted ${\mathbf{L} = \twoel{\mathbf{S}}{\Vector{}{t}{}} \in \mathcal{SE}(3)}$.
Define the system's input ${u = \threeel{\wn}{\tauInp{}{}}{\Vector{}{\mu}{}} \in \mathbb{L} \subseteq \Rn{25}}$.
The core states' input ${\wn = \fourel{\omegaInp{}{}}{\aInp{}{}}{\nuInp{}{}}{1} \in \Rn{10}}$ includes the biased inertial measurements, an additional virtual velocity input ${\nuInp{}{} = \zeronm{3}{1}}$~\cite{Fornasier2022EquivariantBiases}, and a unitary input introduced in~\cite{Delama2025EquivariantApproach}.
${\tauInp{}{} = \threeel{\tauInp{}{\omega}}{\tauInp{}{a}}{\tauInp{}{\nu}} \in \Rn{9}}$ denotes the biases input, and ${\Vector{}{\mu}{} = \twoel{\Vector{}{\mu}{S}}{\Vector{}{\mu}{t}} \in \Rn{6}}$ denotes the calibration states' input.
Finally, by additionally defining the vector ${\gn = \fourel{\zeronm{3}{1}}{-\Vector{}{g}{}}{\zeronm{3}{1}}{1} \in \Rn{10}}$, the continuous-time noise-free system can be represented in compact form as
\begin{equation}
    \dot\xi = f(\xi, u) \quad\Longleftrightarrow\quad 
    \left\{\begin{aligned}
    \dotPose{}{} &= -\gn^\wedge\Pose{}{} + \Pose{}{} (\wn^\wedge - \bias{}{}^\wedge) \\
    \dotbias{}{} &= \tauInp{}{} \\
    \dot{\mathbf{L}} &= \mathbf{L} \Vector{}{\mu}{}^\wedge
    \end{aligned}\right. ,
    \label{eq:system_ct}
\end{equation}
where ${\gn^\wedge, \wn^\wedge \in \g{3}}$, ${\bias{}{}^\wedge \in \setwo{3}}$, and ${\Vector{}{\mu}{}^\wedge \in \se{3}}$.
Strictly speaking, operations such as addition or subtraction between elements of different Lie algebras are not formally defined.
However, since ${\gn^\wedge, \wn^\wedge, \bias{}{}^\wedge \in \Rnm{5}{5}}$ are represented as matrices within a common vector space, such operations are well-defined in that space.
The use of the \emph{wedge} operator here is therefore a matter of notational convenience.
Assuming constant input $u_i$ and bias $\Vector{}{b}{i}$ between consecutive timestamps $t_i$ and $t_{i+1}$, the exact discretization of~\eqref{eq:system_ct} yields the discrete-time formulation ${\xi_{i+1} = F_{\delt}(\xi_i, u_i)}$ for the radar-INS as
\begin{equation}
    \left\{\begin{aligned}
    \Pose{}{i+1} &= \exp(-\gn^\wedge\delt)\Pose{}{i}\exp(({\wn}_i^{\wedge} - \bias{}{i}^{\wedge})\delt) \\
    \bias{}{i+1} &= \bias{}{i} + \tauInp{}{i} \delt \\
    \mathbf{L}_{i+1} &= \mathbf{L}_i \exp(\Vector{}{\mu}{i}^\wedge \delt)
    \end{aligned}\right.,
    \label{eq:system_dt}
\end{equation}
where ${\xi_i = \xi(t_i)}$ and ${u_i = u(t_i)}$ denote respectively the state and the input at the $i$-th timestamp, and ${\delt = t_{i+1} - t_i}$ is the time step.
The time-step dependence of the function is made explicit by the subscript $F_{\delt}$.
Note that the first equation of~\eqref{eq:system_dt} propagates the navigation states using the $\G{3}$ exponential.
To the best of the authors’ knowledge, Galilean symmetry was recently used to model IMU kinematics in preintegration~\cite{Delama2025EquivariantApproach}, but its application to a complete filtering framework is novel.

\subsection{Symmetry of the radar-INS}
\label{sec:symmetry}
The EqF is derived from a \emph{lifted system}, where the original system is represented on a symmetry group associated with its inherent equivariant structure.
This formulation defines the state dynamics in a structured way and leverages the equivariant error with its favorable linearization properties~\cite{vanGoor2022EquivariantEqF}.

Define the symmetry group ${\grpG}$ associated with the radar-INS as ${\grpG \coloneqq \SEtwo{3} \ltimes \setwo{3} \times \SE{3}}$.
Note that this group includes a semi-direct product between $\SEtwo{3}$ and its Lie algebra $\setwo{3}$ (\secref{TangentGroup}).
An element of this group is denoted by ${X = \threeel{D}{\delta}{F} \in \grpG}$, where ${D = \threeel{A}{a}{b} \in \SEtwo{3}}$, ${\delta \in \setwo{3}}$, and ${F = \twoel{E}{f} \in \SE{3}}$.
\begin{lemma}
    The state action ${\phi : \grpG \times \mathcal{M} \rightarrow \mathcal{M}}$ defined as
    \begin{equation}
        \phi(X, \xi) \coloneqq \threeel{\Pose{}{} D}{\AdMsym{D^{-1}}(\bias{}{} - \delta^\vee)}{\PI{\SE{3}\!\!}{D^{-1}}\mathbf{L}F}
        \label{eq:phi}
    \end{equation}
is a transitive and free right group action of $\grpG$ on $\mathcal{M}$.
\end{lemma}
Fix $\mathring{\xi} \in \mathcal{M}$, then since the action is free, the partial map $\phi_{\mathring{\xi}} : \grpG \to \mathcal{M} $ is a diffeomorphism. 
The partial inverse of the state action ${\phi^{-1}_{\mathring{\xi}} : \mathcal{M} \rightarrow \grpG}$ is defined as follows:\vspace{-2mm}
\begin{equation}
    \!\!\!\phi^{-1}_{\mathring{\xi}}\!(\xi)\!=\!\threeel{\!\mathring{\Pose{}{}}^{\!-1}\!\!\Pose{}{}\!}{(\mathring{\bias{}{}} - \AdMsym{\mathring{\Pose{}{}}^{\!-1}\!\!\Pose{}{}}{\bias{}{}})^\wedge\!}{\mathring{\mathbf{L}}^{-1}\PI{\mathcal{SE}(3)}{\mathring{\Pose{}{}}^{\!-1}\!\!\Pose{}{}}\mathbf{L}\!}\!.
    \label{eq:phi_inv}
\end{equation}
\begin{lemma}
    The input action ${\psi : \grpG \times \mathbb{L} \rightarrow \mathbb{L}}$ defined as
    \begin{equation}
        \!\!\psi(\!X, u\!)\!\coloneqq\!\threeel{\!\AdMsym{\PI{\G{3}}{\!D^{-1}\!}}\!(\!\wInp{}{}\!\!-\!\PI{\g{3}}{\delta}^{\!\vee})}{\AdMsym{\!D^{-1}\!}\!\tauInp{}{}}{\AdMsym{\!F^{-1}\!}\!\Vector{}{\mu}{}\!}
        \label{eq:psi}
    \end{equation}
    is a transitive and free right group action of $\grpG$ on $\mathbb{L}$.
\end{lemma}
\begin{theorem}
    The systems~\eqref{eq:system_ct}~\eqref{eq:system_dt} are equivariant under the actions $\phi$~\eqref{eq:phi} and $\psi$~\eqref{eq:psi} of $\grpG$.
    The proof is omitted but can be derived by leveraging the theoretical insights from~\cite[Chapter~5.5]{Fornasier2024EquivariantNavigation}, taking into account that we introduce a novel symmetry structure based on the Galilean group.
    \label{th:eq_sys}
\end{theorem}
The next step for the derivation of the EqF is the definition of the \emph{lifted} system on the symmetry group $\grpG$.
A discrete \emph{lift} for a system ${\xi_{i+1} = F_{\delt}(\xi_i, u_i)}$ is a map ${\Lambda_{\delt} : \mathcal{M} \times \mathbb{L} \rightarrow \grpG}$ subject to ${\phi(\Lambda_{\delt}(\xi, u), \xi) = F_{\delt}(\xi, u)}$, ${\forall \xi \in \mathcal{M}}$ and ${\forall u \in \mathbb{L}}$~\cite{Ge2022EquivariantSystems}.
To improve readability in the following derivations, we drop the subscript indicating the time step dependency of the lift and keep this dependence implicit, i.e., ${\Lambda(\xi, u) \coloneqq \Lambda_{\delt}(\xi, u)}$.
\begin{theorem}
    Define the discrete lift $\Lambda : \mathcal{M} \times \mathbb{L} \rightarrow \grpG$ as
    \begin{equation}
        \Lambda(\xi, u) \coloneqq \threeel{\Lift{D}(\xi, u)}{\Lift{\delta}(\xi, u)}{\Lift{F}(\xi, u)} ,
        \label{eq:lift}
    \end{equation}
    where ${\Lift{D} \in \SEtwo{3}}$, ${\Lift{\delta} \in \setwo{3}}$, and ${\Lift{F} \in \SE{3}}$ are
    \begin{align*}
        \Lift{D}(\xi, u) &= \PI{\SEtwo{3}}{\exp\!\left(-\Adsym{\Pose{}{}^{-1}}{\gn^\wedge}\delt\right) \exp\!\left((\wn^{\wedge} - \bias{}{}^{\wedge})\delt\right)} , \\
        \Lift{\delta}(\xi, u) &= \bias{}{}^{\wedge} - \Adsym{\Lift{D}(\xi, u)}{\bias{}{}^{\wedge} + \tauInp{}{}^{\wedge}\delt} ,\\
        \Lift{F}(\xi, u) &= \Adsym{\mathbf{L}^{-1}}{\PI{\SE{3}}{\Lift{D}(\xi, u)}} \exp(\Vector{}{\mu}{}^\wedge \delt) .
    \end{align*}
    Then, $\Lambda$ is an equivariant lift for the system~\eqref{eq:system_dt} with respect to the symmetry group $\grpG$.
    The proof is omitted due to space limitations, but it draws from~\cite[Chapter~5.5]{Fornasier2024EquivariantNavigation} and~\cite[4.2]{Ge2022EquivariantSystems}.
\end{theorem}
Finally, the \emph{lifted system} evolution on $\grpG$ is given as follows:
\begin{equation}
    X_{i+1} = X_i \Lambda\!\left(\phi_{\mathring{\xi}}(X_i), u_i\right) , \quad X_0 = \phi^{-1}_{\mathring{\xi}}(\xi(t_0)) ,
    \label{eq:lifted_system}
\end{equation}
where ${\mathring{\xi} \in \mathcal{M}}$ is an arbitrarily chosen \emph{state origin}.
Note that we can project elements from the symmetry group $\grpG$ to the manifold $\mathcal{M}$ with ${\xi = \phi_{\mathring{\xi}}(X)}$, and vice versa with ${X = \phi^{-1}_{\mathring{\xi}}(\xi)}$.

\subsection{Equivariant error and propagation matrices}
\label{sec:error}
The symmetry established in the previous subsection enables us to leverage the \emph{equivariant error}~\cite{vanGoor2022EquivariantEqF} to quantify the difference between elements in the homogeneous space and those of the symmetry group.
Specifically, the equivariant error is defined as ${e \coloneqq \phi(\hat{X}^{-1}, \xi) \in \mathcal{M}}$, where ${\hat{X} = \threeel{\hat{D}}{\hat{\delta}}{\hat{F}} \in \grpG}$ denotes the current \emph{state estimate} on the symmetry group, and ${\xi = \threeel{\Pose{}{}}{\bias{}{}}{\mathbf{L}} \in \mathcal{M}}$ represents the \emph{true state} on the manifold.

Define a local parametrization ${\vartheta : \mathcal{M} \rightarrow \R^{24}}$ for the error around the state origin $\mathring{\xi}$ using \emph{normal coordinates}~\cite{vanGoor2022EquivariantEqF} as
\begin{equation}
    \Vector{}{\varepsilon}{} = \vartheta(e) = \log\!\left(\phi_{\mathring{\xi}}^{-1}(e)\right)^{\!\vee}\!\!\! = \log\!\left(\phi_{\mathring{\xi}}^{-1}\!\left(\phi(\hat{X}^{-1}, \xi)\right)\right)^{\!\vee}\!\!\! \in \Rn{24} ,
\end{equation}
where ${\log(\point): \grpG \to \mathfrak{g}}$ denotes the logarithm of the symmetry group $\grpG$.
Consider a \emph{noisy} input ${\tilde{u} = u + \eta \in \mathbb{L}}$ and fix ${\mathring{\xi} = I}$.

The discrete-time filter matrices~\cite{Ge2022EquivariantSystems} are derived by linearizing the error dynamics $\Vector{}{\varepsilon}{i+1} \approx \mathbf{A} \Vector{}{\varepsilon}{i} + \mathbf{B} \Vector{}{\eta}{i}$ about ${\Vector{}{\varepsilon}{} = \Vector{}{0}{} \in \Rn{24}}$ with \scalebox{0.86}{$\mathbf{A}\!\Vector{}{\varepsilon}{}\!=\!\Vector{}{\varepsilon}{} + \Diff{\vartheta}{e}{\mathring{\xi}} \circ \diff{\phi_{\mathring{\xi}}} \circ \Diff{\mathrm{R}_{\mathring{\Lambda}^{-1}}}{Y}{\mathring{\Lambda}} \circ \Diff{\Lambda_{\mathring{u}}}{e}{\mathring{\xi}} \circ \diff{\vartheta^{-1}}\![\Vector{}{\varepsilon}{}]$}, \scalebox{0.86}{$\mathbf{B}\!\Vector{}{\eta}{}\!=\!\Diff{\vartheta}{e}{\mathring{\xi}} \circ \diff{\phi_{\mathring{\xi}}} \circ \Diff{\mathrm{R}_{\mathring{\Lambda}^{-1}}}{Y}{\mathring{\Lambda}} \circ \Diff{\Lambda_{\mathring{\xi}}}{u}{\mathring{u}} \circ \Diff{\psi_{\hat{X}^{-1}}}{u}{\tilde{u}}[\Vector{}{\eta}{}]$}, where ${\mathring{\Lambda} = \Lambda(\mathring{\xi}, \mathring{u})}$ and ${\mathring{u} = \psi(\hat{X}^{-1}\!\!, \tilde{u}) = \threeel{\ringwn}{\ringtauInp{}{}}{\mathring{\Vector{}{\mu}{}}} \in \mathbb{L}}$ is the \emph{origin input}.
The resulting discrete-time \emph{state} matrix $\mathbf{A}$ is
\begin{equation}
    \mathbf{A} = \begin{bmatrix}
        \mathbf{\Gamma} & \mathbf{A}_1 & \zeronm{9}{6} \\
        \zeronm{9}{9} & \mathbf{\Gamma}\mathbf{\Upsilon} & \zeronm{9}{6} \\
        \Xi(\mathbf{\Gamma} - \mathbf{\Gamma}\mathbf{\Upsilon}) & \Xi(\mathbf{A}_1) & \mathbf{A}_2
    \end{bmatrix} \in \Rnm{24}{24} ,
    \label{eq:Adt}
\end{equation}
with ${\mathbf{\Gamma}, \mathbf{\Upsilon}, \mathbf{A}_1 \in \Rnm{9}{9}}$ and ${\mathbf{A}_2 \in \Rnm{6}{6}}$ defined as
\begin{align*}
    &\!\!\mathbf{\Gamma} = \Theta\!\left(\AdMsym{\exp(-\gn^\wedge\delt)}\right), &&\!\!\mathbf{A}_1 = \mathbf{\Gamma} \Theta\!\left(\Jl{\ringwn \delt} \right) \delt , \\
    &\!\!\mathbf{\Upsilon} = \Theta\!\left(\AdMsym{\exp(\ringwn^\wedge\delt)}\right), &&\!\!\mathbf{A}_2 = \AdMsym{\PI{\SE{3}}{\exp(-\gn^\wedge\delt) \exp(\ringwn^\wedge\delt)}} ,
\end{align*}
where ${\Jl{\cdot}}$ and ${\exp(\cdot)}$ denote the $\G{3}$ left Jacobian and exponential map~\cite{Delama2025EquivariantApproach}, respectively.
The linear maps $\Xi$ and $\Theta$ are in~\eqref{eq:Xi} and~\eqref{eq:Theta}.
The discrete-time \emph{input noise} matrix $\mathbf{B}$ is 
\begin{equation}
    \mathbf{B} = \begin{bmatrix}
        \mathbf{B}_1 & \zeronm{9}{9} & \zeronm{9}{6} \\
        \zeronm{9}{10} & \mathbf{\Gamma}\mathbf{\Upsilon} \AdMsym{\hat{D}} \delt & \zeronm{9}{6} \\
        \Xi(\mathbf{B}_1) & \zeronm{6}{9} & \mathbf{B}_2 \\
    \end{bmatrix} \in \Rnm{24}{25} ,
    \label{eq:Bdt}
\end{equation}
where ${\mathbf{B}_1 \in \Rnm{9}{10}}$ and ${\mathbf{B}_2 \in \Rnm{6}{6}}$ are defined as
\begin{align*}
    \mathbf{B}_1 &= \Omega(-\AdMsym{\exp(-\gn^\wedge\delt)} \Jl{\ringwn\delt} \AdMsym{\PI{\G{3}}{\hat{D}}} \delt), \\
    \mathbf{B}_2 &= -\mathbf{A}_2 \Jl{\mathring{\Vector{}{\mu}{}}\delt} \AdMsym{\hat{F}} \delt .
\end{align*}
Note that ${\mathring{\Vector{}{\mu}{}} \in \Rn{6}}$ is defined through the origin input, and $\Jl{\mathring{\Vector{}{\mu}{}}\delt}$ is the $\SE{3}$ left Jacobian, while $\Omega$ is defined in~\eqref{eq:Omega}.

\subsection{Doppler velocity update}
\label{sec:doppler_update}
The radar Doppler velocity measurement ${\vd \in \R}$ is modeled with the function ${\h{v} : \mathcal{M} \times \Rn{3} \times \Rn{3} \to \R}$, defined as
\begin{equation}
    \h{v}(\xi, \pf, \omegaInp{}{}) = -\frac{\pf^\top}{\norm{\pf}} \mathbf{S}^\top \left(\Rot{}{}^\top \vel{}{} + (\omegaInp{}{} - \bias{}{\omega})^\wedge \Vector{}{t}{} \right) ,
    \label{eq:output}
\end{equation}
where ${\pf \in \Rn{3}}$ denotes the respective \emph{feature} position in the radar frame, i.e., the 3D point measured by the radar, and ${\omegaInp{}{} \in \Rn{3}}$ is the biased angular velocity.
To simplify the notation in the following derivation, we treat $\pf$ and $\omegaInp{}{}$ as implicit arguments in the measurement function, i.e., ${\h{v}(\xi) = \h{v}(\xi, \pf, \omegaInp{}{})}$.

The \emph{output} matrix $\Matrix{C}{v}{}$ is derived from the linearized output ${\h{v}(\xi) - \h{v}(\hat{\xi}) \approx \Matrix{C}{v}{}\!\Vector{}{\varepsilon}{}}$, with ${\h{v}(\xi) = \h{v}(\xi, \pf, \omegaInp{}{})}$ as~\eqref{eq:output}, and $\Matrix{C}{v}{}\!\Vector{}{\varepsilon}{} = \Diff{\h{v}}{\xi}{\hat{\xi}} \circ \Diff{\phi_{\hat{X}}}{e}{\mathring{\xi}} \circ \diff{\vartheta^{-1}}[\Vector{}{\varepsilon}{}]$, resulting in
\begin{equation}
    \!\!\!\!\Matrix{C}{v}{}\!\!=\!\begin{bmatrix} \Matrix{C}{v}{1} & \!\!\!\mathbf{\Psi} & \!\!\!\Matrix{C}{v}{2} & \!\!\!-\mathbf{\Psi} (\hat{f} - \hat{b})^\wedge & \!\!\!\zeronm{1}{6} & \!\!\!-\Matrix{C}{v}{1} & \!\!\!-\Matrix{C}{v}{2}
    \end{bmatrix}\!\in\!\Rnm{1}{24} \!\!,
    \label{eq:Cv}
\end{equation}
where ${\mathbf{\Psi} = -\frac{\pf^\top}{\norm{\pf}} \hat{E}^\top \in \Rnm{1}{3}}$ and ${\Matrix{C}{v}{1}, \Matrix{C}{v}{2} \in \Rnm{1}{3}}$ are defined as
\begin{align*}
    &\Matrix{C}{v}{1} = \mathbf{\Psi} \left( \mathring{\omegaInp{}{}}^\wedge \hat{f}^\wedge - (\hat{a} + \mathring{\omegaInp{}{}}^\wedge (\hat{f} - \hat{b}))^\wedge \right) ,
    &&\Matrix{C}{v}{2} = -\mathbf{\Psi} \mathring{\omegaInp{}{}}^\wedge .
\end{align*}
Note that $\mathring{\omegaInp{}{}}$ comes from the origin input ${\ringwn = \fourel{\mathring{\omegaInp{}{}}}{\mathring{\aInp{}{}}}{\mathring{\nuInp{}{}}}{1}}$.

Assuming noisy output measurements, we want to derive the \emph{output noise} matrix $\Matrix{D}{v}{}$ by linearizing the residual with respect to the noise vector.
For the Doppler velocity measurement we have ${\tilde{v}_{\scalebox{0.5}{D}} = \vd + \eta_{\vd}}$.
To better account for the uncertainty in the 3D point measurement, we express it in \emph{spherical coordinates},  following the approach in~\cite{Xu2025IncorporatingSLAM}.
For all ${\Vector{}{x}{} \in \Rn{3}}$ define the map
\begin{align}
    &\varphi : \Rn{3} \to \R \times \mathbb{S}^2, &&\varphi(\Vector{}{x}{}) = \twoel{\kappa}{\Vector{}{\rho}{}} = \twoel{\norm{\Vector{}{x}{}}}{\frac{\Vector{}{x}{}}{\norm{\Vector{}{x}{}}}} ,
\end{align}
where ${\kappa = \norm{\Vector{}{x}{}} \in \R}$, and ${\Vector{}{\rho}{} = \frac{\Vector{}{x}{}}{\norm{\Vector{}{x}{}}} \in \mathbb{S}^2}$ is a unit vector on the \emph{2-sphere}.
The noisy 3D point measurement can be written as:
\begin{equation}
    \curlVector{}{p}{\scalebox{0.5}{$f$}} = \varphi^{-1}(\varphi(\pf) \boxplus_{(\R \times \mathbb{S}^2)} \Vector{}{\eta}{\varphi}) \approx \pf + \Matrix{J}{\varphi}{}(\pf) \Vector{}{\eta}{\varphi} ,
    \label{eq:spherical_coords}
\end{equation}
where ${\Vector{}{\eta}{\varphi} = \twoel{\eta_\kappa}{\Vector{}{\eta}{\rho}} \in \Rn{3}}$ denotes the noise vector in spherical coordinates, and ${\boxplus_{(\R \times \mathbb{S}^2)}}$ denotes the "boxplus" operator for the ${\R \times \mathbb{S}^2}$ manifold~\cite{He2021KalmanManifolds}.
The matrix ${\Matrix{J}{\varphi}{}(\pf)}$ is defined as
\begin{equation}
    \Matrix{J}{\varphi}{}(\pf) = \begin{bmatrix} \Vector{}{\rho}{} & -\kappa \Vector{}{\rho}{}^\wedge \Matrix{N}{\rho}{} \end{bmatrix} \in \Rnm{3}{3} ,
\end{equation}
with ${\kappa = \norm{\pf} \in \R}$, ${\Vector{}{\rho}{} = \frac{\pf}{\norm{\pf}} \in \mathbb{S}^2}$, and ${\Matrix{N}{\rho}{} \in \Rnm{3}{2}}$ defined as
\begin{equation}
    \Matrix{N}{\rho}{} = \exp\!\left(\!\left(\frac{\Vector{}{e}{3}^\wedge \Vector{}{\rho}{}}{\norm{\Vector{}{e}{3}^\wedge \Vector{}{\rho}{}}} \atantwo\!\left( \norm{\Vector{}{e}{3}^\wedge \Vector{}{\rho}{}}, \Vector{}{e}{3}^\top \Vector{}{\rho}{} \right)\!\right)^{\!\!\wedge} \right) \begin{bmatrix} \Vector{}{e}{1} & \Vector{}{e}{2} \end{bmatrix} .
\end{equation}
Here, ${\Vector{}{e}{1}, \Vector{}{e}{2}, \Vector{}{e}{3} \in \Rn{3}}$ denote the standard basis of $\Rn{3}$.

Finally, the \emph{output noise} matrix $\Matrix{D}{v}{}$ is derived by linearizing the output equation as ${\tilde{v}_{\scalebox{0.5}{D}} - \h{v}(\xi, \curlVector{}{p}{\scalebox{0.5}{$f$}}, \tilde{u}) \approx \Matrix{D}{v}{}\Vector{}{\zeta}{}^{\scalebox{0.6}{$v$}}}$, with respect to the noise vector ${\Vector{}{\zeta}{}^{\scalebox{0.6}{$v$}} = \threeel{\Vector{}{\eta}{\omega}}{\Vector{}{\eta}{\varphi}}{\eta_{\vd}} \in \Rn{7}}$, which results in
\begin{equation}
    \Matrix{D}{v}{}= \begin{bmatrix} \mathbf{\Psi} (\hat{f} - \hat{b})^\wedge \hat{A} & \Matrix{D}{v}{1} & 1
    \end{bmatrix} \in \Rnm{1}{7} ,
    \label{eq:Dv}
\end{equation}
where ${\Matrix{D}{v}{1} \in \Rnm{1}{3}}$ is defined as
\begin{equation*}
    \Matrix{D}{v}{1} = \frac{1}{\norm{\pf}} \left( \hat{E}^\top (\hat{a} + \mathring{\omegaInp{}{}}^\wedge (\hat{f} - \hat{b})) \right)^\top \left( \eyen{3} - \frac{\pf \pf^\top}{\norm{\pf}^2} \right) \Matrix{J}{\varphi}{}(\pf) .
\end{equation*}
We have derived all the necessary components for designing the EqF for RIO, and the implementation is detailed in~\algoref{eqf}.

%% file: sections/multistate.tex
\section{Multi-state Constraint}
\label{sec:msc}
To account for geometric constraints introduced by detecting the same radar point feature from multiple poses, the EqF-RIO presented in~\secref{filter} is extended with \emph{multi-state constraint} updates, similarly to~\cite{Fornasier2023MSCEqF:Navigation}.
Point associations between radar scans are determined with the learning-based approach in~\cite{DLmatching}.

\subsection{State augmentation and propagation matrices}
\label{sec:augment}
The original system state $\xi = \threeel{\Pose{}{}}{\bias{}{}}{\mathbf{L}} \in \mathcal{M}$, as defined in~\secref{ins}, is augmented with a sliding window of the most recent $k$ radar poses $\mathbf{P}_1, ..., \mathbf{P}_k \in \mathcal{SE}(3)$ with zero dynamics, each corresponding to a timestamp where one or multiple 3D feature positions $\pf$ were observed and tracked.
The \emph{augmented} state in the manifold is denoted $\xi^a = \fourel{\xi}{\mathbf{P}_1}{\ldots}{\mathbf{P}_k} \in \mathcal{M} \times \mathcal{SE}(3)^k$.
Define the lifted system state $X^a = \fourel{X}{F_1}{\ldots}{F_k} \in \grpG \times \SE{3}^k$, where ${X = \threeel{D}{\delta}{F} \in \grpG}$ is defined in~\secref{symmetry}.
As each element ${F_i \in \{ F_1, ..., F_k \}}$ represents a static pose from a prior timestamp, the associated state action and discrete lift are
\begin{align}
    \stateaction{F_i}(X, \xi) &= \mathbf{P}_i F_i \in \mathcal{SE}(3) , &&\Lift{F_i}(\xi, u) = I \in \SE{3} .
\end{align}
The augmented state and input matrices for $k$ pose clones are
\begin{align}
    \Matrix{A}{a}{} &= \begin{bmatrix}
        \mathbf{A} & \zeronm{24}{(6k)} \\
        \zeronm{(6k)}{24} & \eyen{(6k)}
    \end{bmatrix} \in \Rnm{(24+6k)}{(24+6k)} , \\
    \Matrix{B}{a}{} &= \begin{bmatrix} \Matrix{B}{}{} \\ \zeronm{(6k)}{25} \end{bmatrix} \in \Rnm{(24+6k)}{25}
\end{align}
where ${\mathbf{A} \in \Rnm{24}{24}}$ is defined in~\eqref{eq:Adt} and ${\mathbf{B} \in \Rnm{24}{25}}$ in~\eqref{eq:Bdt}.

\subsection{Multi-state constraint update}
\label{sec:msc_update}
To leverage temporal consistency in feature observations, we introduce a measurement model that constrains the current system state using past radar observations.
Specifically, consider a single 3D feature point $\pf$ observed at the current time and at a previous time $t_i$, associated with the past radar pose $\mathbf{P}_i$.
Define the measurement function ${\h{p} : \mathcal{M} \times \Rn{3} \to \R}$ as
\begin{equation}
    \h{p}(\xi, \pfi) = \norm{\left( \PI{\mathcal{SE}(3)}{\mathbf{T}} \mathbf{L} \right)^{-1} \mathbf{P}_i * \pfi} ,
    \label{eq:hp}
\end{equation}
where $\pfi$ denotes the 3D point measurement at time $t_i$ and the operator ${* : \mathcal{SE}(3) \times \Rn{3} \to \Rn{3}}$ is given by ${\mathbf{P} * \Vector{}{x}{} = \mathbf{R} \Vector{}{x}{} + \Vector{}{p}{}}$ where ${\mathbf{P} = \twoel{\mathbf{R}}{\Vector{}{p}{}} \in \mathcal{SE}(3)}$ and ${\Vector{}{x}{} \in \Rn{3}}$.
Since radar features are often noisy and do not maintain consistent identities across frames, comparing full 3D positions directly can be unreliable.
Instead, we use the Euclidean norm of the transformed point~\cite{Michalczyk2022Tightly-CoupledOdometry}, which yields a simpler and more robust scalar error term that emphasizes spatial consistency while being less sensitive to feature orientation and noise.

Without loss of generality, consider the case of a single pose clone $\mathbf{P}_i$.
The \emph{output} matrix $\Matrix{C}{p}{}$ is derived from the linearized output ${\h{p}(\xi) - \h{p}(\hat{\xi}) \approx \Matrix{C}{p}{}\!\Vector{}{\varepsilon}{}}$, with ${\h{p}(\xi) = \h{p}(\xi, \pfi)}$ as~\eqref{eq:hp}, and $\Matrix{C}{p}{}\!\Vector{}{\varepsilon}{} = \Diff{\h{p}}{\xi}{\hat{\xi}} \circ \Diff{\phi_{\hat{X}}}{e}{\mathring{\xi}} \circ \diff{\vartheta^{-1}}[\Vector{}{\varepsilon}{}]$, resulting in
\begin{equation}
    \Matrix{C}{p}{} = \begin{bmatrix} \zeronm{1}{18} & \Matrix{C}{p}{1} & -\mathbf{H} & -\Matrix{C}{p}{1} & \mathbf{H}
    \end{bmatrix} \in \Rnm{1}{30},
    \label{eq:Cp}
\end{equation}
where ${\Matrix{C}{p}{1}, \mathbf{H} \in \Rnm{1}{3}}$ are defined as
\begin{align*}
    \Matrix{C}{p}{1} &= \mathbf{H} (\hat{F}_i * \pfi)^\wedge, &&\mathbf{H} = \frac{(\hat{F}^{-1}\hat{F}_i * \pfi)^\top}{\norm{\hat{F}^{-1}\hat{F}_i * \pfi}} \hat{E}^\top .
\end{align*}

The \emph{output noise} matrix $\Matrix{D}{p}{}$ for the multi-state constraint update is derived by linearizing ${\norm{\curlVector{}{p}{\scalebox{0.5}{$f$}}} - \h{p}(\xi, \curlVector{}{p}{\scalebox{0.5}{$f$}i}) \approx \Matrix{D}{p}{} \Vector{}{\zeta}{}^{\scalebox{0.6}{$p$}}}$ with respect to ${\Vector{}{\zeta}{}^{\scalebox{0.6}{$p$}} = \twoel{\Vector{}{\eta}{\varphi}}{\Vector{}{\eta}{\varphi_i}} \in \Rn{6}}$.
Here, ${\eta_\varphi, \eta_{\varphi_i} \in \Rn{3}}$ denote the noise of the 3D point at current time and at past time $t_i$, given in spherical coordinates~\eqref{eq:spherical_coords}.
The resulting matrix is
\begin{equation}
    \Matrix{D}{p}{} = \begin{bmatrix} 1 & 0 & 0 & - \mathbf{H} \hat{E}_i \Matrix{J}{\varphi}{}(\pfi) \end{bmatrix} \in \Rnm{1}{6} .
\end{equation}

\input{algorithms/eqf}

%% file: algorithms/eqf.tex
\begin{algorithm}[t]
\caption{Equivariant Filter for Radar-Inertial Odometry}
\begin{flushleft}
    \textbf{Parameters:} ${\mathring{\xi} = \threeel{\eyen{5}}{\zeronm{9}{1}}{\eyen{4}}}$, $\xi_{init}$, ${\mathbf{\Sigma}_{init}}$, $\mathbf{Q}$, $\mathbf{R}$ \\
    \textbf{Initialization:} ${\hat{X} = \phi^{-1}_{\mathring{\xi}}(\xi_{init})}$, ${\hat{\mathbf{\Sigma}} = \mathbf{\Sigma}_{init}}$ \\
    \textbf{Measurements:} IMU meas. $\twoel{\curlomegaInp{}{}}{\curlaInp{}{}}$, radar meas. $\twoel{\curlVector{}{p}{f}}{\tilde{v}_{D}}$ \\
    \textbf{Output:} ${\hat{X} \rightarrow \hat{\xi} = \phi_{\mathring{\xi}}(\hat{X})}$ and $\hat{\mathbf{\Sigma}}$ at time $t_{curr}$ 
\end{flushleft}
\vspace{-3mm}
\hrulefill\\
\textsc{Propagation}: require an input meas. ${\tilde{u} = \threeel{\curlwn}{\tauInp{}{}}{\Vector{}{\mu}{}}}$
\begin{algorithmic}[1]
    \State ${\delt \gets t_{curr} - t_{prev}}$
    \State ${\hat{\mathbf{\Sigma}} \gets \mathbf{A} \hat{\mathbf{\Sigma}} \mathbf{A}^\top + \frac{1}{\delt} \mathbf{B} \mathbf{Q} \mathbf{B}^\top }$ \hfill \eqref{eq:Adt}~\eqref{eq:Bdt}
    \State ${\hat{X} \gets \hat{X} \Lambda(\phi_{\mathring{\xi}}(\hat{X}), \tilde{u})}$ \hfill \eqref{eq:phi}~\eqref{eq:lift}~\eqref{eq:lifted_system}
\end{algorithmic}
\textsc{Update}: require an output meas. ${\tilde{y} = \twoel{\curlVector{}{p}{\scalebox{0.5}{$f$}}}{\tilde{v}_{\scalebox{0.5}{D}}}}$
\begin{algorithmic}[1]
    \State ${\Vector{}{r}{} \gets \tilde{v}_{\scalebox{0.5}{D}} - \h{v}(\phi_{\mathring{\xi}}(\hat{X}),\curlVector{}{p}{\scalebox{0.5}{$f$}},\curlwInp{}{})}$ \hfill \eqref{eq:output}
    \State ${\mathbf{K} \gets \hat{\mathbf{\Sigma}}\mathbf{C}^\top (\mathbf{C} \hat{\mathbf{\Sigma}} \mathbf{C}^\top + \mathbf{D} \mathbf{R} \mathbf{D}^\top)^{-1}}$ \hfill \eqref{eq:Cv}~\eqref{eq:Dv}
    \State ${\Delta \gets \diff{\phi_{\mathring{\xi}}}^\dag \circ \diff{\vartheta^{-1}} \circ \mathbf{K} \Vector{}{r}{}}$ \hfill {\scriptsize{$^\star$Note that ${\diff{\phi_{\mathring{\xi}}}^\dag \circ \diff{\vartheta^{-1}}[\point] = (\point)^\wedge}$}}
    \State ${\hat{\mathbf{\Sigma}} \gets (\eyen{24} - \mathbf{K}\mathbf{C})\hat{\mathbf{\Sigma}}}$
    \State ${\hat{X} \gets \exp(\Delta) \hat{X}}$
\end{algorithmic}
\label{alg:eqf}
\end{algorithm}

%% file: sections/results.tex
\section{Experiments and Results}
\label{sed:results}
\input{figures/ntnu}
To validate the proposed EqF-RIO, we ran experiments on two real-world datasets, \texttt{arena} and \texttt{ardea}, each collected using a different UAV platform shown in~\figref{UAVs}.
Metadata for each dataset sequence is summarized in~\tabref{metadata}.
The \texttt{arena} dataset was collected using a custom-built UAV equipped with a VectorNav VN-100 IMU, a Texas Instruments IWR6843AOP radar (configured according to~\cite{doer2021radar}), and an Ouster OS0-128 LiDAR, all synchronized by a microcontroller.
% These sensors were time-synchronized by an onboard microcontroller.
The UAV was manually flown in a test arena following three different trajectories ranging from 76\,m to 155\,m, as illustrated in~\figref{ntnu_plot}.
Ground-truth poses were obtained by fusing LiDAR and IMU measurements using the LiDAR–inertial odometry framework of \cite{Nissov2024DegradationOdometry}, with radar factors disabled.
This provides a reliable reference trajectory, given the geometrically well-conditioned structure of the test environment.
The \texttt{ardea} dataset was captured with the ARDEA-X UAV~\cite{ardea, Michalczyk2024Tightly-CoupledOdometry} and contains five trajectories ranging from 12\,m to 38\,m, with some sequences flown manually and others following pre-planned waypoints.

We evaluate the performance of EqF-RIO using standard metrics.
The \emph{Absolute Pose Error} (APE) measures the difference between estimated and ground truth poses in the global frame, providing an overall assessment of the localization accuracy.
For a trajectory with $M$ poses, let ${\mathbf{R}_i, \Vector{}{p}{i}}$ denote the ground truth rotation and position at time $t_i$, and ${\hat{\mathbf{R}}_i, \hatVector{}{p}{i}}$ the corresponding filter estimates.
The APE can be calculated separately for \emph{rotation} and \emph{translation} as follows:
\begin{equation}
    \!\!\mathrm{APE}^{\mathrm{R}}_i\!= \log(\mathbf{R}_i^\top \hat{\mathbf{R}}_i)^\vee \in \Rn{3}\!, \quad
    \mathrm{APE}^{\mathrm{T}}_i\!= \mathbf{R}_i^\top(\hatVector{}{p}{i} - \Vector{}{p}{i}) \in \Rn{3}\!.
\end{equation}
To summarize rotation and translation APE as single scalar values, we compute the \emph{Root Mean Square Error} (RMSE) as
\begin{equation}
    \mathrm{RMSE} = \sqrt{\frac{1}{M} \sum_{i=0}^{M-1} \mathrm{APE}_i^2} \in \R,
\end{equation}
To quantify filter consistency, we compute the pose \emph{Average Normalized Estimation Error Squared} (ANEES) as
\begin{equation}
    \mathrm{ANEES} = \frac{1}{6M} \sum_{i=0}^{M-1} \Vector{}{\varepsilon}{\scalebox{0.5}{P}i}^\top \mathbf{\Sigma}_{\scalebox{0.5}{P}i}^{-1} \Vector{}{\varepsilon}{\scalebox{0.5}{P}i},
\end{equation}
where ${\Vector{}{\varepsilon}{\scalebox{0.5}{P}i} \in \mathbb{R}^6}$ is the pose estimation error in local coordinates at time $t_i$ and ${\mathbf{\Sigma}_{\scalebox{0.5}{P}i} \in \Rnm{6}{6}}$ is the corresponding pose covariance.

\input{tables/experiments}
\input{figures/ardea}
Table~\ref{tab:metadata} summarizes EqF-RIO performance on both \texttt{arena} and \texttt{ardea} datasets with correct initial radar-IMU extrinsic calibration.
Results are grouped into two categories: "Doppler Only", which uses only Doppler velocity updates (\secref{doppler_update}), and "Doppler \& Matches", which additionally uses multi-state constraint updates (\secref{msc_update}) using 3D point correspondences obtained with the matching method in~\cite{DLmatching}.
For each sequence, we report both translation and rotation APE RMSE as well as the position and yaw drift, defined as the final error normalized by trajectory length.
For the \texttt{arena} dataset, \figref{ntnu_plot} provides a visual comparison of estimated and ground truth trajectories and their APE.
While raw RMSE values provide a quantitative indication of estimation accuracy and are useful for comparing different estimators on the same sequence, they naturally grow with trajectory length since position and yaw remain unobservable without global information.
Drift is therefore the most informative metric.
Results show that EqF-RIO achieves low positional drift, typically below 1\% of the traveled distance.
Exploiting matches in multi-state constraint updates further improves robustness by enforcing spatial consistency across different poses, mitigating extrinsic calibration errors and reducing drift accumulation in longer or more complex trajectories.
However, matched features are sparser and noisier than Doppler velocities, so they may increase the RMSE in particular sequences, e.g., \texttt{ardea-5}.
\input{tables/test}
\input{figures/error}
To evaluate robustness and basin-of-attraction properties, we introduce an initial radar–IMU calibration error $e_\angle$ defined as
\begin{equation}
    e_\angle \coloneqq \norm{\log(\mathbf{S}_0^\top \hat{\mathbf{S}}_0)^\vee}
    \label{eq:e_angle}
\end{equation}
where ${\mathbf{S}_0 \in \mathcal{SO}(3)}$ denotes the initial radar–IMU rotation matrix (\secref{ins}).
We compared our EqF-RIO against a state-of-the-art EKF-RIO~\cite{Michalczyk2022Tightly-CoupledOdometry}, which shares the same filter updates, including multi-state constraints.
All experiments were performed under identical initial conditions and noise parameters to ensure a fair comparison.
To separate the effect of the equivariant formulation from that of the radar noise model, we consider two variants of our approach: EqF* employs the same radar point noise model as the EKF in~\cite{Michalczyk2022Tightly-CoupledOdometry}, i.e., \emph{without spherical-coordinates modeling}, while EqF denotes the full equivariant implementation using spherical coordinates~\eqref{eq:spherical_coords}.
Table~\ref{tab:comparison} summarizes these experiments for increasing $e_\angle$, reporting translation and rotation RMSE, pose ANEES, and extrinsics convergence.
Convergence is classified as follows: \emph{converged} (green), i.e., the radar-IMU rotation error norm decreases to a small (5$^\circ$), bounded residual, \emph{partial} (yellow), i.e., the error shows a clear decreasing trend but does not reach full convergence within the available trajectory, and \emph{fail} (red), i.e., the error does not decrease at all, leading to filter divergence.
For small $e_\angle$ both filters achieve good accuracy and consistency, although EqF-RIO performs slightly better, likely due to its improved measurement noise modeling and better error linearization.
At 10$^\circ$ error, EKF-RIO already exhibit partial convergence where EqF-RIO performs reliably (\figref{ardea1_plot_10}).
With substantial initial error, EKF-RIO consistently fails, whereas EqF-RIO maintains strong performance as highlighted in Figures~\ref{fig:UAVs},~\ref{fig:ardea1_plot_80}, and~\ref{fig:convergence}.
At 80$^\circ$ error, EqF-RIO reaches full convergence in all cases with the only exception of \texttt{ardea-5}, where the trajectory geometry provides limited excitation as motion occurs primarily along a single axis with minimal rotations, limiting observability and making estimation particularly challenging, with partial convergence at 10$^\circ$ error for both filters.
This case highlights that evaluating only raw RMSE values can be misleading.
In fact, RMSE values may appear relatively small even if the filter fails to converge, underlining the importance of considering also convergence and ANEES.
Larger initial errors still show the advantage of EqF-RIO, though it is being pushed to its limits.
At 100$^\circ$ error, our method achieves either full or partial convergence in most cases, whereas at 120$^\circ$ error it fully converges only in one case and otherwise exhibits partial convergence or failure.
The difference between EqF$^*$ and EqF suggests that most of the performance gain in terms of basin-of-attraction comes from the equivariant symmetry rather than the improved noise modeling~\eqref{eq:spherical_coords}.
Finally, the ANEES values further underline the difference between the two approaches.
When EKF-RIO fails, ANEES grows far above unity, indicating strong inconsistency as the covariance no longer reflects the true estimation error.
In contrast, EqF-RIO maintains much better ANEES across all experiments, confirming that the improved linearization properties of the equivariant error preserve consistency even when the calibration states are far from their true values.

%% file: figures/ntnu.tex
\begin{figure*}[t]
    \centering
    % Column 1
    \subfloat[\texttt{arena-1}]{%
        \begin{minipage}[b]{0.32\textwidth}
            \centering
            % Top row: two images side by side
            \begin{minipage}[b]{0.49\textwidth}
                \includegraphics[width=\textwidth]{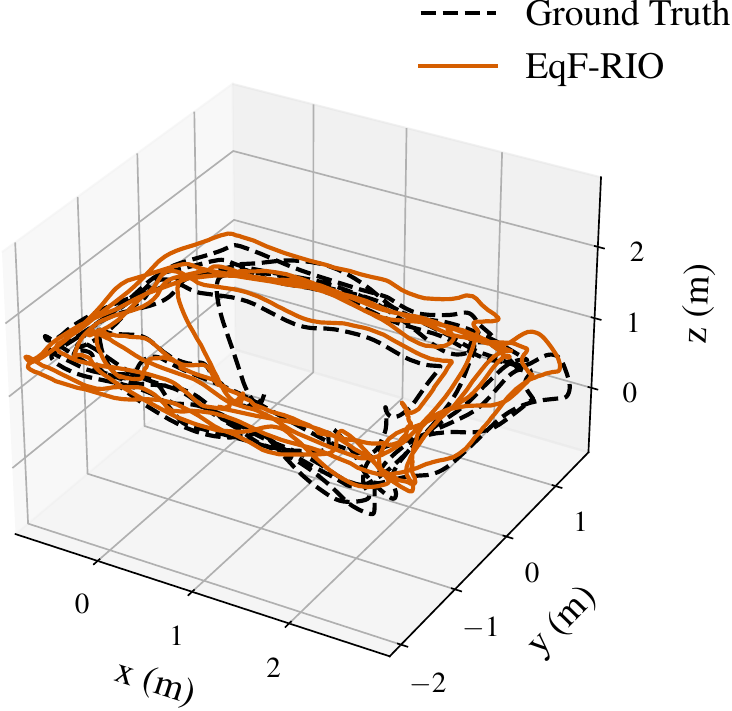}
            \end{minipage}%
            \hfill
            \begin{minipage}[b]{0.49\textwidth}
                \includegraphics[width=\textwidth]{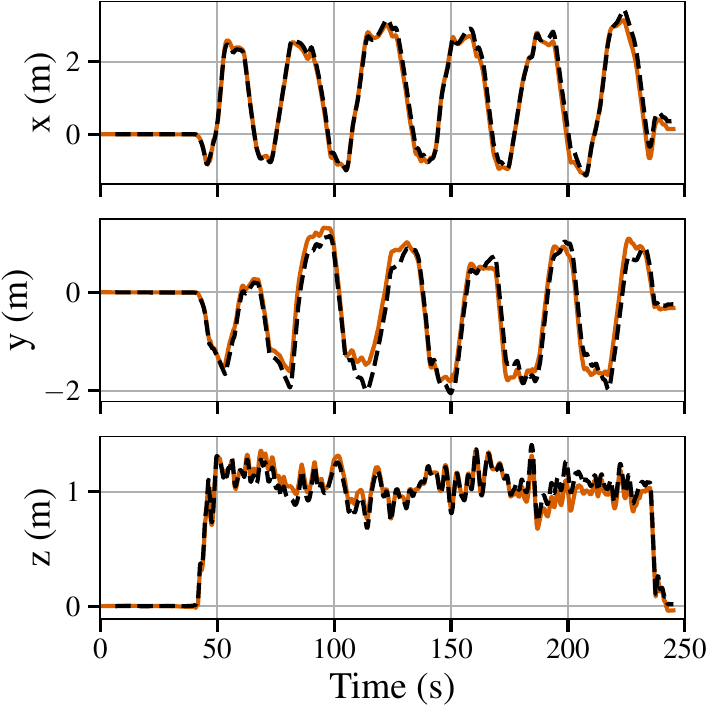}
            \end{minipage}
            \par\vspace{0.5mm} % <--- forces vertical space
            \includegraphics[width=\textwidth]{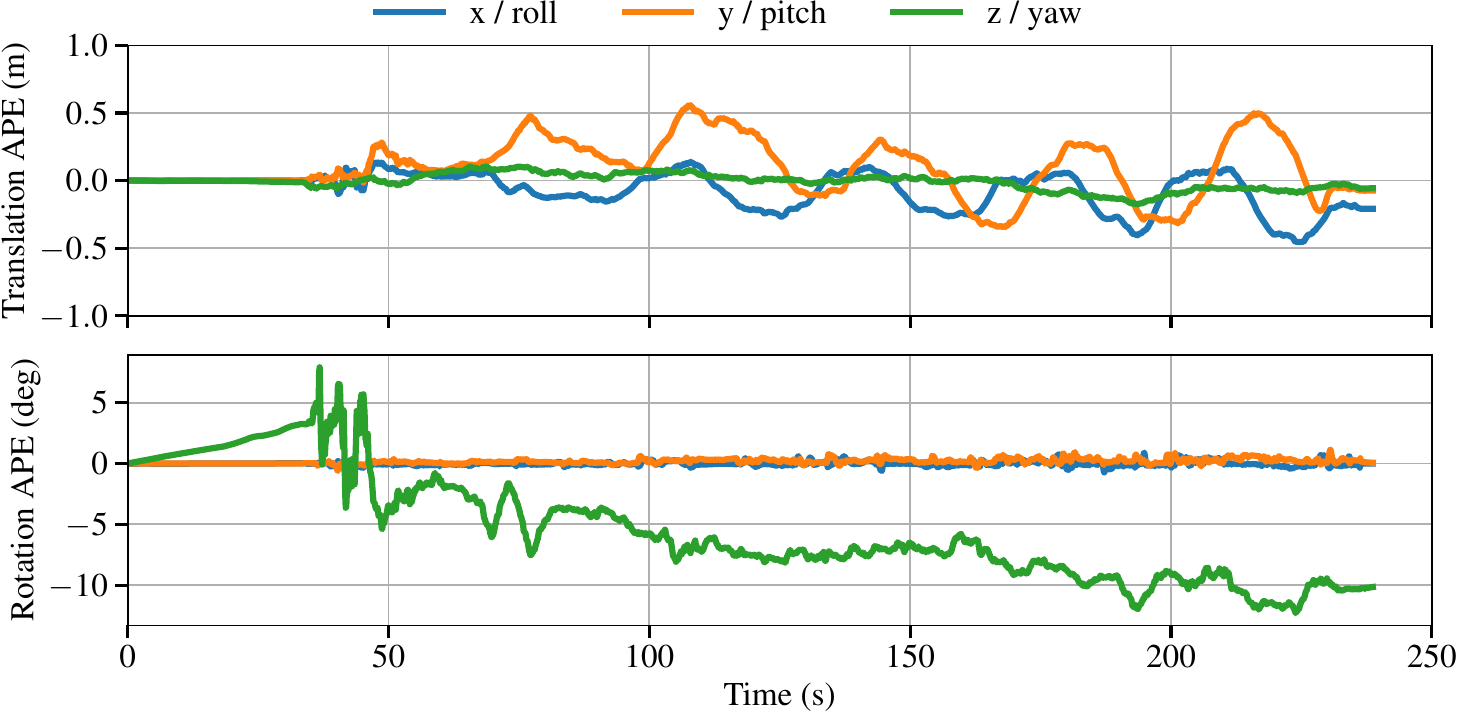}
        \vspace{-5.5mm}
        \end{minipage}%
    }
    \hfill
    % Column 2
    \subfloat[\texttt{arena-2}]{%
        \begin{minipage}[b]{0.32\textwidth}
            \centering
            \begin{minipage}[b]{0.49\textwidth}
                \includegraphics[width=\textwidth]{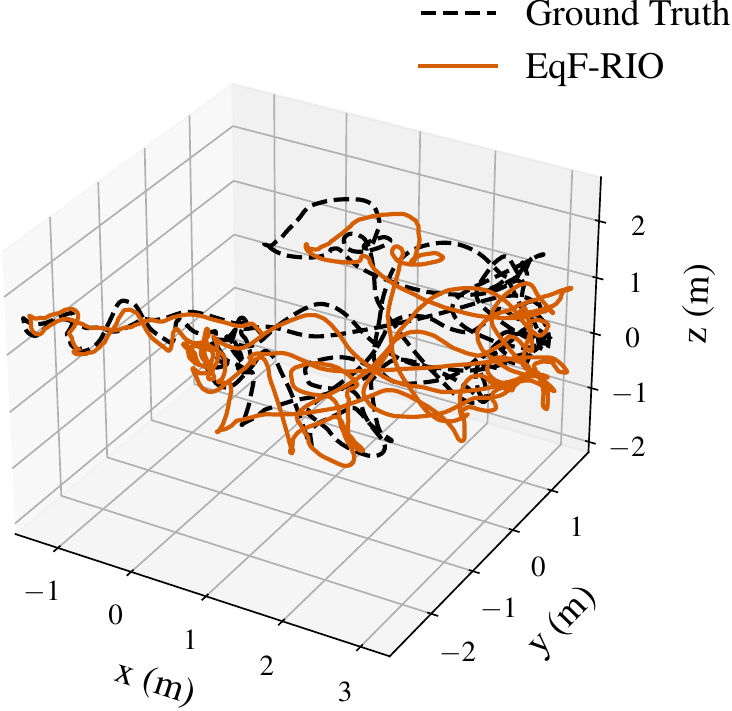}
            \end{minipage}%
            \hfill
            \begin{minipage}[b]{0.49\textwidth}
                \includegraphics[width=\textwidth]{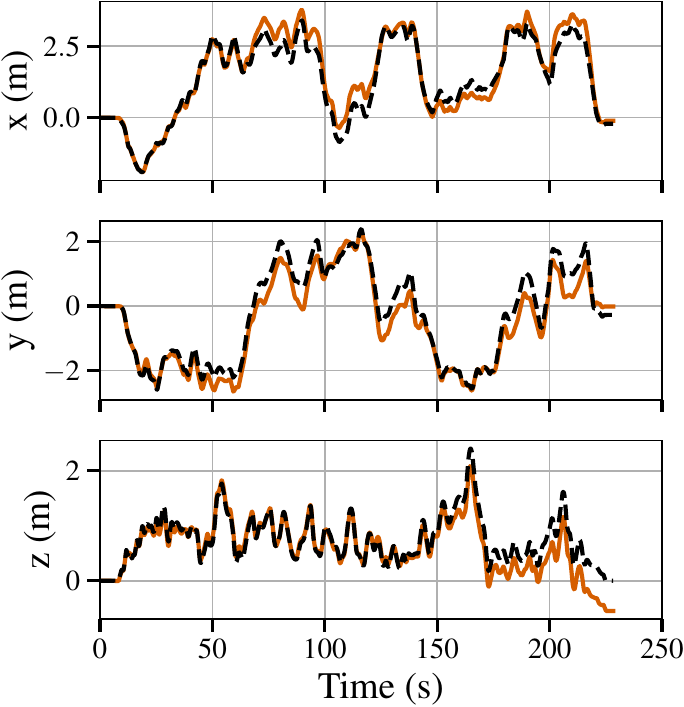}
            \end{minipage}
            \par\vspace{0.5mm} % vertical space
            \includegraphics[width=\textwidth]{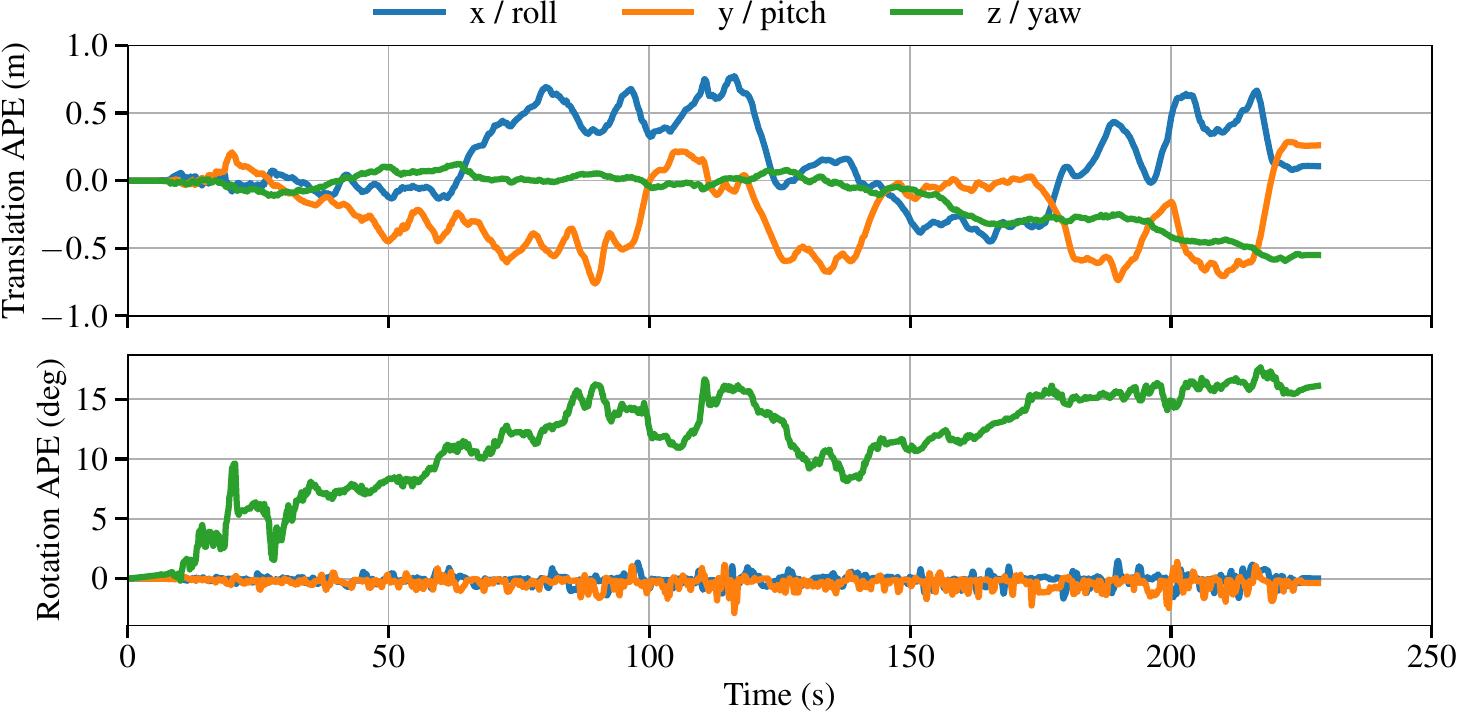}
        \vspace{-5.5mm}
        \end{minipage}%
    }
    \hfill
    % Column 3
    \subfloat[\texttt{arena-3}]{%
        \begin{minipage}[b]{0.32\textwidth}
            \centering
            \begin{minipage}[b]{0.49\textwidth}
                \includegraphics[width=\textwidth]{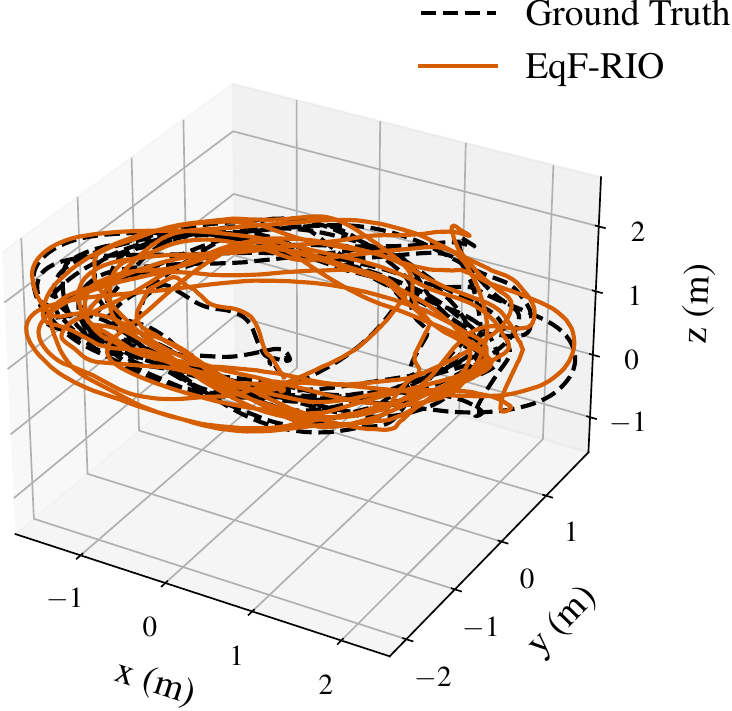}
            \end{minipage}%
            \hfill
            \begin{minipage}[b]{0.49\textwidth}
                \includegraphics[width=\textwidth]{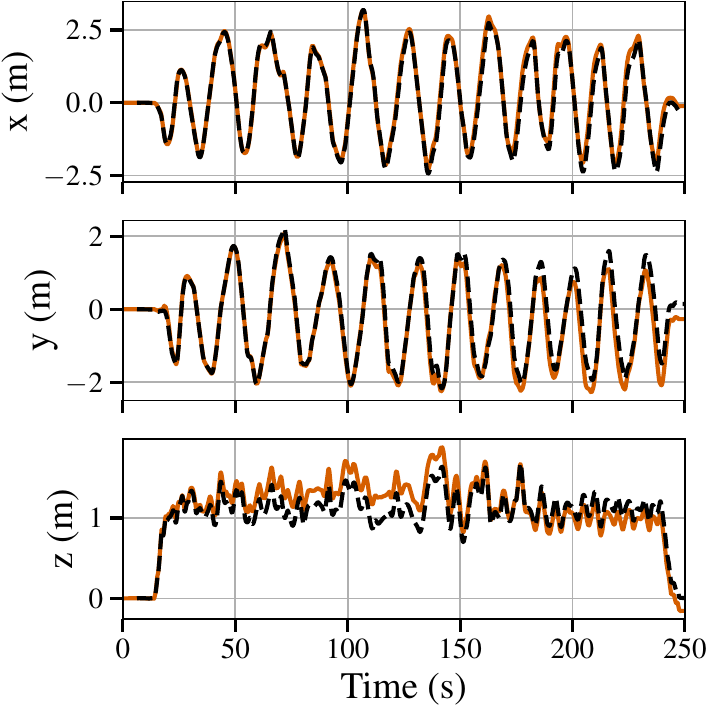}
            \end{minipage}
            \par\vspace{0.5mm} % vertical space
            \includegraphics[width=\textwidth]{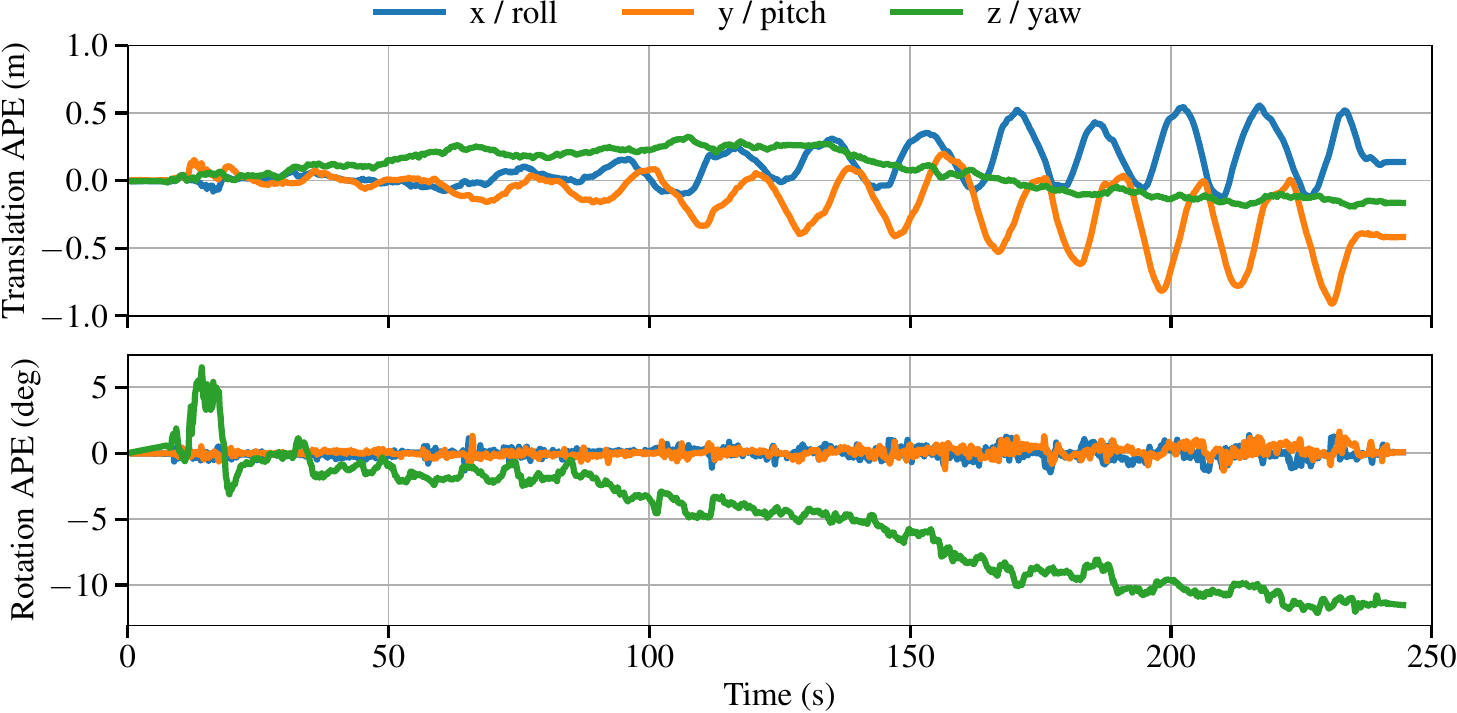}
        \vspace{-5.5mm}
        \end{minipage}%
    }
    \vspace{-1mm}
    \caption{Evaluation of EqF-RIO on the three \texttt{arena} sequences using only Doppler velocity updates. Each sequence shows the 3D trajectory (top-left), position components (top-right), and translation and rotation APE (bottom). Quantitative APE RMSE and drift metrics are reported in \tabref{metadata}. The APE plots clearly illustrate the gradual accumulation of position and yaw drift when no global information is available.}
    \vspace{-5mm}
    \label{fig:ntnu_plot}
\end{figure*}

%% file: tables/experiments.tex
\begin{table}[t]
\vspace{-4mm}
\centering
\setlength{\tabcolsep}{5pt}
\caption{Real-world datasets overview and EqF-RIO results}\vspace{-3mm}
\label{tab:metadata}
\scriptsize
\newcommand{\rotation}{90}
\begin{tabular}{@{}l c c c cc cc@{}}
\toprule
& \multicolumn{3}{c}{Experiments metadata}
  & \multicolumn{2}{c}{APE RMSE} 
  & \multicolumn{2}{c}{Drift} \\
\cmidrule(lr){2-4}\cmidrule(lr){5-6}\cmidrule(lr){7-8}
 & \makecell{Dataset\\Sequence}
 & \makecell{Len.\\(m)} 
 & \makecell{$v_{\mathrm{max}}$\\(m/s)}
 & \makecell{Translation\\(m)} 
 & \makecell{Rotation\\(deg)} 
 & \makecell{Position\\(cm/m)} 
 & \makecell{Yaw\\(deg/m)} \\
\midrule
\multirow{8}{*}{\rotatebox{\rotation}{\makecell{Doppler\\Only}}}
 & \texttt{arena-1} & 76  & 0.88 & 0.28 & 6.99 & 0.3 & 0.13 \\
 & \texttt{arena-2} & 82  & 0.99 & 0.54 & 12.15 & 0.8 & 0.20 \\
 & \texttt{arena-3} & 155 & 1.24 & 0.38 & 6.63 & 0.3 & 0.07 \\
 \cmidrule[.1pt](lr){2-8}
 & \texttt{ardea-1} & 12 & 0.66 & 0.09 & 1.28 & 0.7 & 0.07 \\
 & \texttt{ardea-2} & 20 & 0.69 & 0.25 & 1.88 & 1.0 & 0.15 \\
 & \texttt{ardea-3} & 38 & 0.90 & 0.26 & 2.18 & 0.8 & 0.13 \\
 & \texttt{ardea-4} & 23 & 0.93 & 0.19 & 2.87 & 2.7 & 0.57 \\
 & \texttt{ardea-5} & 14 & 0.76 & 0.14 & 5.99 & 1.4 & 0.93 \\
\midrule
\multirow{5}{*}{\rotatebox{\rotation}{\makecell{Doppler\\\& Matches}}}
 & \texttt{ardea-1} & 12 & 0.66 & 0.07 & 2.86 & 0.4 & 0.33 \\
 & \texttt{ardea-2} & 20 & 0.69 & 0.21 & 5.31 & 0.8 & 0.42 \\
 & \texttt{ardea-3} & 38 & 0.90 & 0.20 & 6.13 & 0.6 & 0.39 \\
 & \texttt{ardea-4} & 23 & 0.93 & 0.19 & 2.90 & 2.5 & 0.51 \\
 & \texttt{ardea-5} & 14 & 0.76 & 0.40 & 5.20 & 7.1 & 0.69 \\
\bottomrule
\multicolumn{8}{c}{\tiny Table results use correct initial radar-IMU calibration. Position drift (cm/m) equals drift percentage (\%).}
\vspace{-6mm}
\end{tabular}
\end{table}

%% file: figures/ardea.tex
\begin{figure*}[!t]
    \centering
    % Column 1
    \subfloat[\texttt{ardea-1} ($e_{\angle} = 10^\circ$)]{%
        \begin{minipage}[b]{0.48\textwidth}
            \centering
            % Top row
            \begin{minipage}[b]{0.49\textwidth}
                \includegraphics[width=\textwidth]{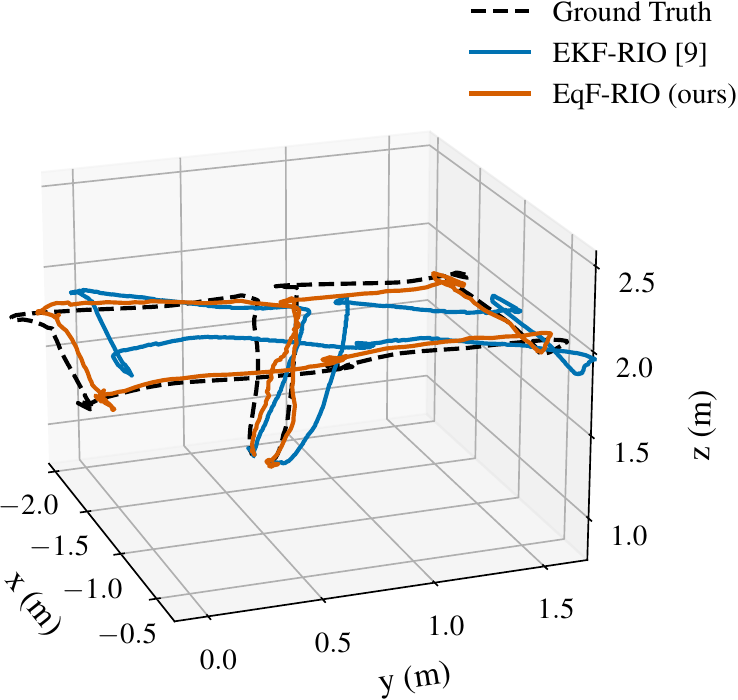}
            \end{minipage}%
            \hfill
            \begin{minipage}[b]{0.49\textwidth}
                \includegraphics[width=\textwidth]{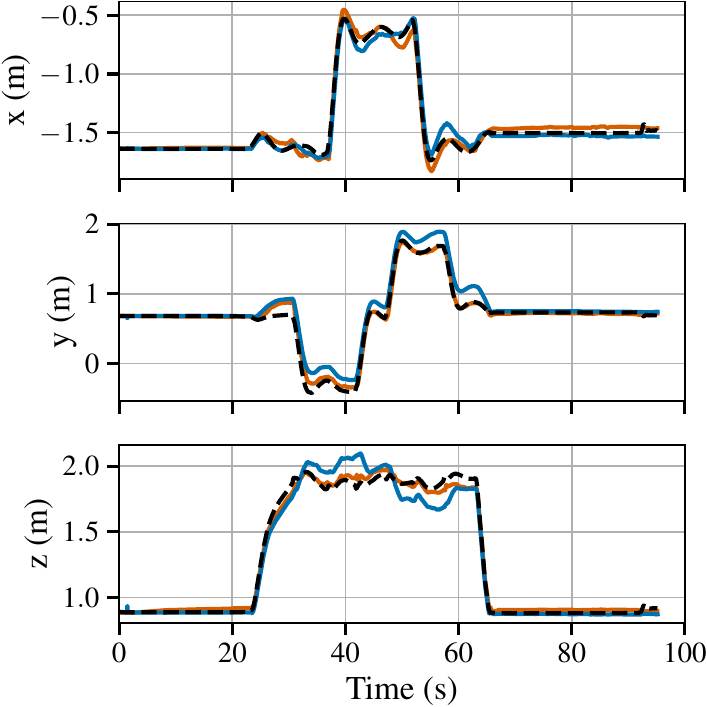}
            \end{minipage}
            
            % New row (third figure)
            \vspace{0.5ex}
            \includegraphics[width=\textwidth]{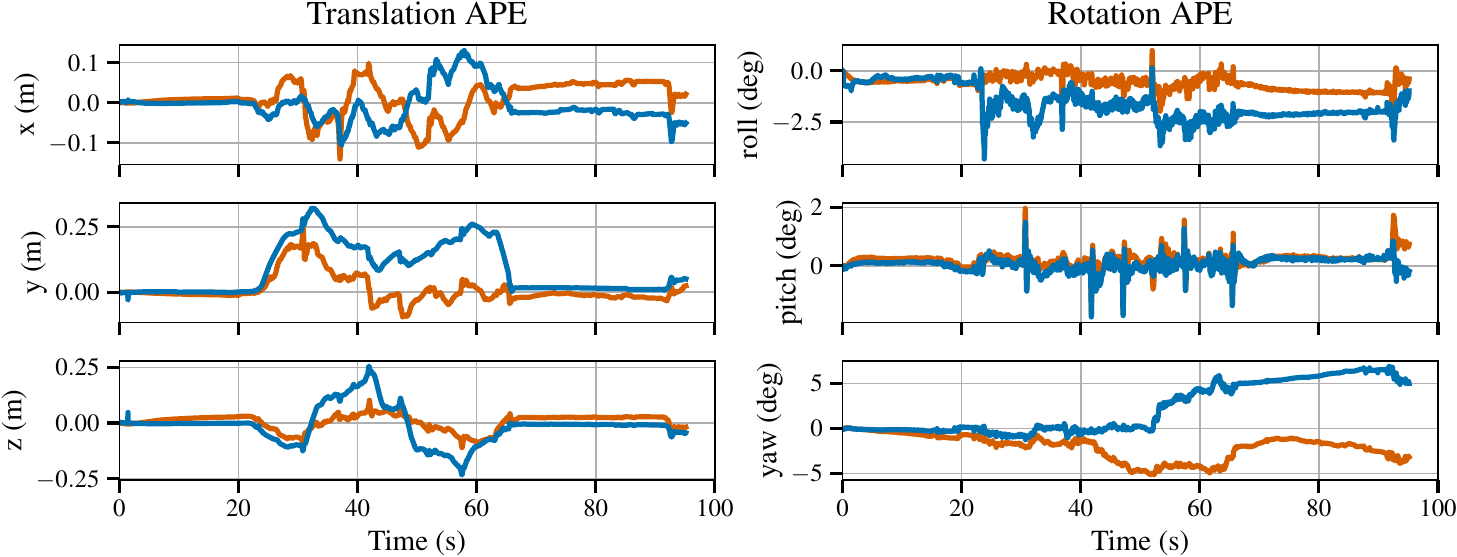}\label{fig:ardea1_plot_10}
        \vspace{-5mm}
        \end{minipage}%
    }
    \hfill
    % Column 2
    \subfloat[\texttt{ardea-1} ($e_{\angle} = 80^\circ$)]{%
        \begin{minipage}[b]{0.48\textwidth}
            \centering
            % Top row
            \begin{minipage}[b]{0.49\textwidth}
                \includegraphics[width=\textwidth]{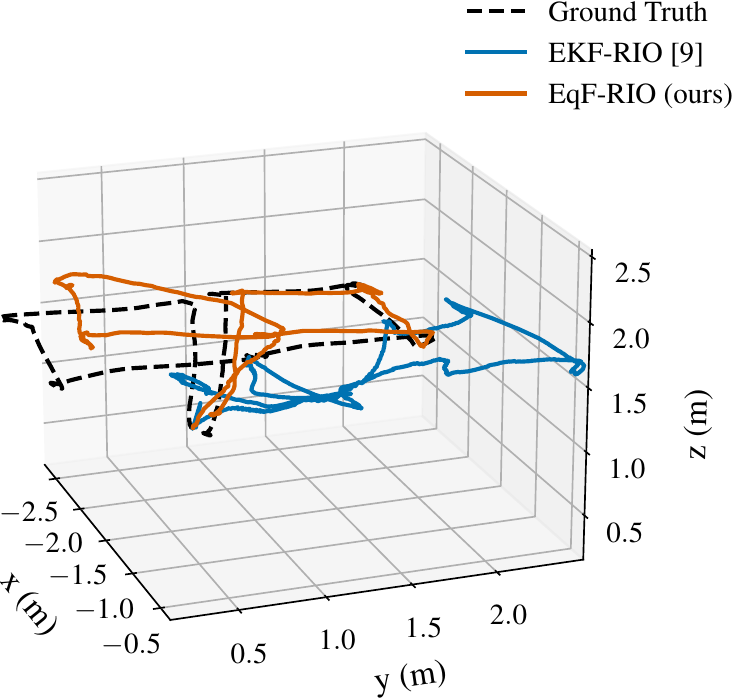}
            \end{minipage}%
            \hfill
            \begin{minipage}[b]{0.49\textwidth}
                \includegraphics[width=\textwidth]{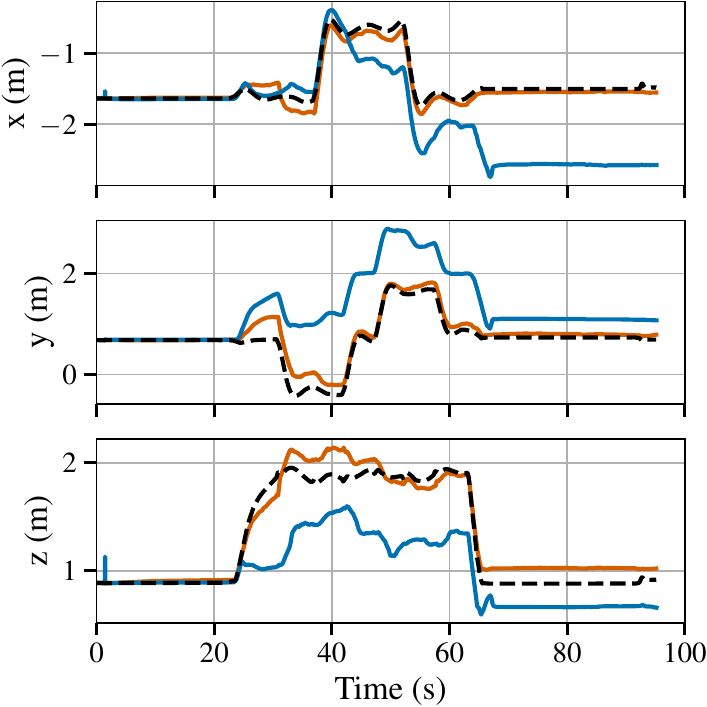}
            \end{minipage}
            
            % New row (third figure)
            \vspace{0.5ex}
            \includegraphics[width=\textwidth]{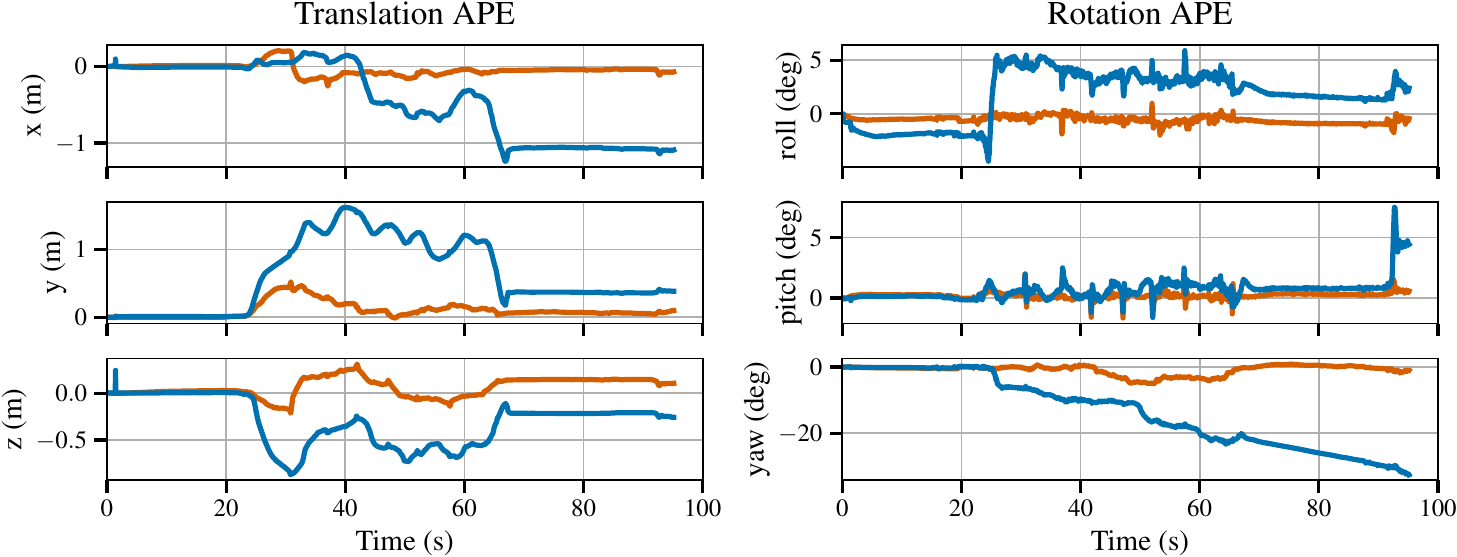}\label{fig:ardea1_plot_80}
        \vspace{-5mm}
        \end{minipage}%
    }
    \vspace{-1mm}
    \caption{Comparison between EKF-RIO~\cite{Michalczyk2022Tightly-CoupledOdometry} (blue) and EqF-RIO (orange) on the \texttt{ardea-1} sequence for increasing initial radar–IMU orientation calibration errors $e_\angle$. Each run shows four plots: 3D trajectories (top-left), position components (top-right), translation APE (bottom-left), and rotation APE (bottom-right). With a $10^\circ$ error, both filters track the ground truth trajectory, but EKF-RIO achieves only partial convergence of the calibration states, whereas EqF-RIO fully converges and achieves a better APE. With $80^\circ$ error, the EKF diverges while EqF-RIO maintains stable convergence, recovering correct extrinsic calibration after a short transient phase and providing accurate pose estimates. This highlights the robustness of the proposed filter under severe calibration errors.}
    \label{fig:ardea1_plot}
    \vspace{-5mm}
\end{figure*}

%% file: tables/test.tex
\begin{table}[t]
\vspace{-1mm}
\centering
\caption{EKF-RIO~\cite{Michalczyk2022Tightly-CoupledOdometry} vs. EqF-RIO: APE RMSE and Pose ANEES.}\vspace{-3mm}
\label{tab:comparison}
\scriptsize
\setlength{\tabcolsep}{3.5pt}
\newcommand{\rotation}{90}
\begin{tabular}{@{\extracolsep{\fill}}c r | ccc | ccc | ccc |@{}}
\toprule
 \multicolumn{2}{c}{} & \multicolumn{3}{c}{Transl. RMSE (m)} & \multicolumn{3}{c}{Rot. RMSE (deg)} & \multicolumn{3}{c}{Pose ANEES} \\
 \cmidrule(lr){3-5}\cmidrule(lr){6-8}\cmidrule(lr){9-11}
  & $e_{\!\angle}$ & EKF & EqF$^*$ & EqF & EKF & EqF$^*$ & EqF & EKF & EqF$^*$ & EqF \\
\midrule
\multirow{5}{*}{\rotatebox{\rotation}{\texttt{ardea-1}}}
 & 0$^\circ$   & \cellcolor{lightgreen} 0.09 & \cellcolor{lightgreen} 0.08 & \cellcolor{lightgreen} \textbf{0.07} & \cellcolor{lightgreen} 3.81 & \cellcolor{lightgreen} 3.19 & \cellcolor{lightgreen} \textbf{2.86} & \cellcolor{lightgreen} 6.74 & \cellcolor{lightgreen} 1.94 & \cellcolor{lightgreen} \textbf{1.86} \\
 & 10$^\circ$ & \cellcolor{lightyellow} 0.16 & \cellcolor{lightgreen} 0.09 & \cellcolor{lightgreen} \textbf{0.08} & \cellcolor{lightyellow} 3.98 & \cellcolor{lightgreen} 3.05 & \cellcolor{lightgreen} \textbf{2.59} & \cellcolor{lightyellow} 8.92 & \cellcolor{lightgreen} 2.01 & \cellcolor{lightgreen} \textbf{1.93} \\
 & 80$^\circ$  & \cellcolor{lightred} 1.10 & \cellcolor{lightgreen} \textbf{0.19} & \cellcolor{lightgreen} 0.21 & \cellcolor{lightred} 17.54 & \cellcolor{lightgreen} 2.77 & \cellcolor{lightgreen} \textbf{1.90} & \cellcolor{lightred} 68.68 & \cellcolor{lightgreen} 2.24 & \cellcolor{lightgreen} \textbf{2.21} \\
 & 100$^\circ$ & \cellcolor{lightred} 1.08 & \cellcolor{lightgreen} 0.26 & \cellcolor{lightgreen} \textbf{0.26} & \cellcolor{lightred} 6.19 & \cellcolor{lightgreen} 4.27 & \cellcolor{lightgreen} \textbf{2.53} & \cellcolor{lightred} 1.3e2 & \cellcolor{lightgreen} 2.67 & \cellcolor{lightgreen} \textbf{2.50} \\
 & 120$^\circ$ & \cellcolor{lightred} 1.22 & \cellcolor{lightgreen} 0.37 & \cellcolor{lightgreen} \textbf{0.34} & \cellcolor{lightred} 11.24 & \cellcolor{lightgreen} 8.26 & \cellcolor{lightgreen} \textbf{5.20} & \cellcolor{lightred} 1.5e2 & \cellcolor{lightgreen} 3.51 & \cellcolor{lightgreen} \textbf{3.29} \\
\midrule
\multirow{5}{*}{\rotatebox{\rotation}{\texttt{ardea-2}}}
 & 0$^\circ$   & \cellcolor{lightgreen} 0.22 & \cellcolor{lightgreen} 0.24 & \cellcolor{lightgreen} \textbf{0.21} & \cellcolor{lightgreen} \textbf{1.51} & \cellcolor{lightgreen} 2.82 & \cellcolor{lightgreen} 5.31 & \cellcolor{lightgreen} \textbf{3.49} & \cellcolor{lightgreen} 4.36 & \cellcolor{lightgreen} 4.46 \\
 & 10$^\circ$ & \cellcolor{lightgreen} 0.24 & \cellcolor{lightgreen} 0.23 & \cellcolor{lightgreen} \textbf{0.20} & \cellcolor{lightgreen} \textbf{1.60} & \cellcolor{lightgreen} 2.95 & \cellcolor{lightgreen} 4.98 & \cellcolor{lightgreen} \textbf{3.92} & \cellcolor{lightgreen} 4.20 & \cellcolor{lightgreen} 4.32 \\
 & 80$^\circ$  & \cellcolor{lightred} 6.41 & \cellcolor{lightgreen} 0.30 & \cellcolor{lightgreen} \textbf{0.27} & \cellcolor{lightred} 10.79 & \cellcolor{lightgreen} \textbf{3.14} & \cellcolor{lightgreen} 3.89 & \cellcolor{lightred} 5.3e2 & \cellcolor{lightgreen} 5.20 & \cellcolor{lightgreen} \textbf{4.95} \\
 & 100$^\circ$ & \cellcolor{lightred} 10.88 & \cellcolor{lightyellow} 1.25 & \cellcolor{lightyellow} \textbf{0.95} & \cellcolor{lightred} 5.94 & \cellcolor{lightyellow} \textbf{1.69} & \cellcolor{lightyellow} 1.76 & \cellcolor{lightred} 1.2e3 & \cellcolor{lightyellow} 27.42 & \cellcolor{lightyellow} \textbf{21.07} \\
 & 120$^\circ$ & \cellcolor{lightred} 1.5e2 & \cellcolor{lightyellow} 1.49 & \cellcolor{lightyellow} \textbf{1.10} & \cellcolor{lightred} 7.32 & \cellcolor{lightyellow} 4.23 & \cellcolor{lightyellow} \textbf{3.49} & \cellcolor{lightred} 1.2e4 & \cellcolor{lightyellow} 37.25 & \cellcolor{lightyellow} \textbf{25.56} \\
\midrule
\multirow{5}{*}{\rotatebox{\rotation}{\texttt{ardea-3}}}
 & 0$^\circ$   & \cellcolor{lightgreen} 0.28 & \cellcolor{lightgreen} 0.20 & \cellcolor{lightgreen} \textbf{0.20} & \cellcolor{lightgreen} \textbf{3.69} & \cellcolor{lightgreen} 9.67 & \cellcolor{lightgreen} 6.13 & \cellcolor{lightgreen} 32.62 & \cellcolor{lightgreen} 5.91 & \cellcolor{lightgreen} \textbf{5.71} \\
 & 10$^\circ$ & \cellcolor{lightgreen} 0.30 & \cellcolor{lightgreen} 0.21 & \cellcolor{lightgreen} \textbf{0.19} & \cellcolor{lightgreen} \textbf{5.85} & \cellcolor{lightgreen} 7.23 & \cellcolor{lightgreen} 6.05 & \cellcolor{lightgreen} 34.80 & \cellcolor{lightgreen} 5.88 & \cellcolor{lightgreen} \textbf{5.74} \\
 & 80$^\circ$  & \cellcolor{lightred} 2.6e2 & \cellcolor{lightgreen} 0.19 & \cellcolor{lightgreen} \textbf{0.12} & \cellcolor{lightred} 10.44 & \cellcolor{lightgreen} 7.90 & \cellcolor{lightgreen} \textbf{2.55} & \cellcolor{lightred} 1.3e4 & \cellcolor{lightgreen} 5.84 & \cellcolor{lightgreen} \textbf{5.17} \\
 & 100$^\circ$ & \cellcolor{lightred} 5.88 & \cellcolor{lightred} 4.42 & \cellcolor{lightyellow} \textbf{1.00} & \cellcolor{lightred} 10.46 & \cellcolor{lightred} 7.02 & \cellcolor{lightyellow} \textbf{4.53} & \cellcolor{lightred} 4.4e2 & \cellcolor{lightred} 2.0e2 & \cellcolor{lightyellow} \textbf{23.46} \\
 & 120$^\circ$ & \cellcolor{lightred} 45.29 & \cellcolor{lightred} 6.93 & \cellcolor{lightred} \textbf{4.20} & \cellcolor{lightred} \textbf{10.09} & \cellcolor{lightred} 15.49 & \cellcolor{lightred} 21.96 & \cellcolor{lightred} 3.3e3 & \cellcolor{lightred} 3.0e2 & \cellcolor{lightred} \textbf{1.3e2} \\
\midrule
\multirow{5}{*}{\rotatebox{\rotation}{\texttt{ardea-4}}}
 & 0$^\circ$   & \cellcolor{lightgreen} 0.34 & \cellcolor{lightgreen} \textbf{0.18} & \cellcolor{lightgreen} 0.19 & \cellcolor{lightgreen} 10.74 & \cellcolor{lightgreen} \textbf{2.40} & \cellcolor{lightgreen} 2.90 & \cellcolor{lightgreen} 31.14 & \cellcolor{lightgreen} \textbf{3.79} & \cellcolor{lightgreen} 3.89 \\
 & 10$^\circ$ & \cellcolor{lightgreen} 0.39 & \cellcolor{lightgreen} \textbf{0.19} & \cellcolor{lightgreen} 0.20 & \cellcolor{lightgreen} 10.75 & \cellcolor{lightgreen} 2.17 & \cellcolor{lightgreen} \textbf{2.08} & \cellcolor{lightgreen} 31.26 & \cellcolor{lightgreen} \textbf{3.81} & \cellcolor{lightgreen} 4.07 \\
 & 80$^\circ$  & \cellcolor{lightred} 1.52 & \cellcolor{lightgreen} \textbf{0.32} & \cellcolor{lightgreen} 0.34 & \cellcolor{lightred} 14.10 & \cellcolor{lightgreen} 1.86 & \cellcolor{lightgreen} \textbf{1.56} & \cellcolor{lightred} 55.42 & \cellcolor{lightgreen} \textbf{4.23} & \cellcolor{lightgreen} 4.37 \\
 & 100$^\circ$ & \cellcolor{lightred} 7.68 & \cellcolor{lightgreen} 0.43 & \cellcolor{lightgreen} \textbf{0.41} & \cellcolor{lightred} 13.04 & \cellcolor{lightgreen} 2.24 & \cellcolor{lightgreen} \textbf{1.65} & \cellcolor{lightred} 2.9e2 & \cellcolor{lightgreen} \textbf{4.65} & \cellcolor{lightgreen} 4.69 \\
 & 120$^\circ$ & \cellcolor{lightred} 17.08 & \cellcolor{lightyellow} 0.69 & \cellcolor{lightyellow} \textbf{0.63} & \cellcolor{lightred} 15.96 & \cellcolor{lightyellow} 4.57 & \cellcolor{lightyellow} \textbf{3.47} & \cellcolor{lightred} 8.2e2 & \cellcolor{lightyellow} 6.40 & \cellcolor{lightyellow} \textbf{6.06} \\
\midrule
\multirow{5}{*}{\rotatebox{\rotation}{\texttt{ardea-5}}}
 & 0$^\circ$   & \cellcolor{lightgreen} \textbf{0.20} & \cellcolor{lightgreen} 0.33 & \cellcolor{lightgreen} 0.40 & \cellcolor{lightgreen} 13.07 & \cellcolor{lightgreen} \textbf{3.55} & \cellcolor{lightgreen} 5.20 & \cellcolor{lightgreen} 6.12 & \cellcolor{lightgreen} \textbf{3.14} & \cellcolor{lightgreen} 3.16 \\
 & 10$^\circ$  & \cellcolor{lightyellow} \textbf{0.29} & \cellcolor{lightyellow} 0.45 & \cellcolor{lightyellow} 0.47 & \cellcolor{lightyellow} 13.55 & \cellcolor{lightyellow} \textbf{3.62} & \cellcolor{lightyellow} 5.30 & \cellcolor{lightyellow} 6.45 & \cellcolor{lightyellow} \textbf{3.18} & \cellcolor{lightyellow} 3.22 \\
 & 80$^\circ$  & \cellcolor{lightred} \textbf{0.59} & \cellcolor{lightred} 0.73 & \cellcolor{lightred} 0.66 & \cellcolor{lightred} 36.46 & \cellcolor{lightred} \textbf{6.67} & \cellcolor{lightred} 9.12 & \cellcolor{lightred} 49.92 & \cellcolor{lightred} \textbf{3.84} & \cellcolor{lightred} 3.84 \\
 & 100$^\circ$ & \cellcolor{lightred} \textbf{0.62} & \cellcolor{lightred} 0.83 & \cellcolor{lightred} 0.79 & \cellcolor{lightred} 14.86 & \cellcolor{lightred} \textbf{4.83} & \cellcolor{lightred} 5.98 & \cellcolor{lightred} 14.39 & \cellcolor{lightred} \textbf{4.32} & \cellcolor{lightred} 4.54 \\
 & 120$^\circ$ & \cellcolor{lightred} \textbf{0.74} & \cellcolor{lightred} 0.89 & \cellcolor{lightred} 0.92 & \cellcolor{lightred} 33.24 & \cellcolor{lightred} \textbf{4.29} & \cellcolor{lightred} 4.29 & \cellcolor{lightred} 47.88 & \cellcolor{lightred} \textbf{4.52} & \cellcolor{lightred} 5.02 \\
\bottomrule
\multicolumn{11}{c}{\tiny 
$e_{\scalebox{0.7}{$\!\angle$}}$: initial rot. error~\eqref{eq:e_angle}.\quad
$^*$: w/o spherical coords~\eqref{eq:spherical_coords}.\quad
\textbf{Bold}: best result.\quad
{\color{lightgreen}\rule{1.5mm}{1.5mm}}: converged.\quad
{\color{lightyellow}\rule{1.5mm}{1.5mm}}: partial.\quad
{\color{lightred}\rule{1.5mm}{1.5mm}}: fail.
}
\vspace{-7mm}
\end{tabular}
\end{table}

%% file: figures/error.tex
\begin{figure}[t]
    \centering
    \includegraphics[width=\linewidth]{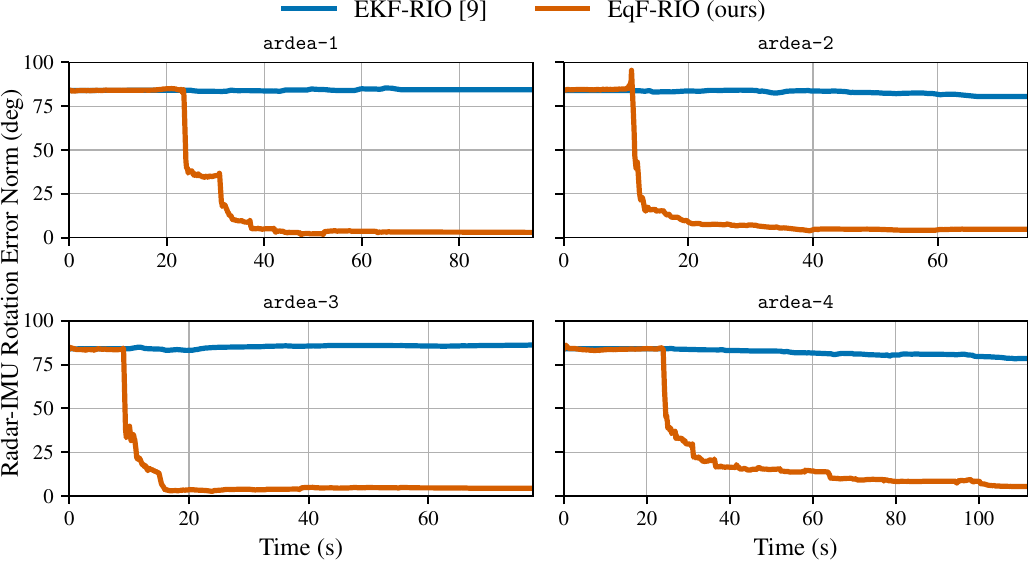}
    \vspace{-7mm}
    \caption{Extrinsic calibration convergence from an initial $80^\circ$ radar–IMU rotation error across four ardea sequences. EqF-RIO (orange) rapidly converges to the correct calibration in all cases, whereas EKF-RIO (blue) consistently fails to recover. The top-left plot (\texttt{ardea-1}) shows a particular transient phase before full convergence, likely due to the trajectory geometry (\figref{ardea1_plot_80}). In fact, motion along different axes provides independent observability of the extrinsic calibration states, so information arrives asynchronously across dimensions. Overall, EqF-RIO demonstrates robust and reliable convergence even under large initialization errors, whereas EKF-RIO fails. Note that the final error does not reach exactly zero because the reference radar-IMU extrinsic calibration is obtained manually and is therefore only approximate.}
    \label{fig:convergence}
    \vspace{-6mm}
\end{figure}

%% file: sections/conclusion.tex
\section{Conclusion}
We presented EqF-RIO, a novel discrete-time EqF for radar-aided INSs that fuses radar Doppler velocities and point-cloud measurements while jointly estimating radar-IMU extrinsic calibration and supporting multi-state constraint updates.
By leveraging an equivariant symmetry and properly accounting for sensor noise and model uncertainties, EqF-RIO preserves consistency and improves robustness.
Analytical discrete-time linearization matrices ensure computational efficiency by eliminating the need for matrix exponentials in filter propagation, while multi-state constraints enforce spatial consistency across subsequent robot poses.
Real-world experiments on two UAV datasets demonstrate state-of-the-art accuracy with positional drift typically below 1\% of the traveled distance under precise calibration, and improved reliability maintaining stable convergence even when recovering from large initial radar–IMU orientation errors of up to 120$^\circ$, outperforming conventional EKF-based methods.
We open-source the evaluation code.

%% file: bibliography/EqF.bib
@article{Fornasier2024AnSystem,
    title = {{An Equivariant Approach to Robust State Estimation for the ArduPilot Autopilot System}},
    year = {2024},
    journal = {Proceedings - IEEE International Conference on Robotics and Automation},
    author = {Fornasier, Alessandro and Ge, Yixiao and Van Goor, Pieter and Scheiber, Martin and Tridgell, Andrew and Mahony, Robert and Weiss, Stephan},
    pages = {11956--11962},
    publisher = {Institute of Electrical and Electronics Engineers Inc.},
    isbn = {9798350384574},
    doi = {10.1109/ICRA57147.2024.10611108},
    issn = {10504729}
}

@article{vanGoor2022EquivariantEqF,
    title = {{Equivariant Filter (EqF)}},
    year = {2022},
    journal = {IEEE Transactions on Automatic Control},
    author = {van Goor, Pieter and Hamel, Tarek and Mahony, Robert},
    number = {6},
    month = {6},
    pages = {3501--3512},
    volume = {68},
    publisher = {Institute of Electrical and Electronics Engineers Inc.},
    doi = {10.1109/TAC.2022.3194094},
    issn = {15582523},
    arxivId = {2010.14666}
}

@article{VanGoor2020EquivariantSpaces,
    title = {{Equivariant Filter (EqF): A General Filter Design for Systems on Homogeneous Spaces}},
    year = {2020},
    journal = {Proceedings of the IEEE Conference on Decision and Control},
    author = {Van Goor, Pieter and Hamel, Tarek and Mahony, Robert},
    number = {Cdc},
    pages = {5401--5408},
    volume = {2020-Decem},
    isbn = {9781728174471},
    doi = {10.1109/CDC42340.2020.9303813},
    issn = {07431546}
}

@article{Ge2022EquivariantSystems,
    title = {{Equivariant Filter Design for Discrete-time Systems}},
    year = {2022},
    journal = {Proceedings of the IEEE Conference on Decision and Control},
    author = {Ge, Yixiao and Van Goor, Pieter and Mahony, Robert},
    pages = {1243--1250},
    volume = {2022-December},
    publisher = {Institute of Electrical and Electronics Engineers Inc.},
    isbn = {9781665467612},
    doi = {10.1109/CDC51059.2022.9992342},
    issn = {25762370}
}

@article{Fornasier2022EquivariantBiases,
    title = {{Equivariant Filter Design for Inertial Navigation Systems with Input Measurement Biases}},
    year = {2022},
    journal = {2022 International Conference on Robotics and Automation (ICRA)},
    author = {Fornasier, Alessandro and Ng, Yonhon and Mahony, Robert and Weiss, Stephan},
    month = {5},
    pages = {4333--4339},
    publisher = {IEEE},
    url = {https://ieeexplore.ieee.org/document/9811778/},
    isbn = {978-1-7281-9681-7},
    doi = {10.1109/ICRA46639.2022.9811778}
}

@article{Fornasier2024EquivariantNavigation,
    title = {{Equivariant Symmetries for Aided Inertial Navigation}},
    year = {2024},
    journal = {arXiv:2407.14297},
    author = {Fornasier, Alessandro},
    month = {7},
    url = {https://arxiv.org/abs/2407.14297v1},
    arxivId = {2407.14297}
}

@article{Fornasier2023MSCEqF:Navigation,
    title = {{MSCEqF: A Multi State Constraint Equivariant Filter for Vision-aided Inertial Navigation}},
    year = {2023},
    journal = {IEEE Robotics and Automation Letters},
    author = {Fornasier, Alessandro and Goor, Pieter van and Allak, Eren and Mahony, Robert and Weiss, Stephan},
    month = {1},
    publisher = {Institute of Electrical and Electronics Engineers Inc.},
    doi = {10.1109/LRA.2023.3335775},
    issn = {23773766}
}

@article{Fornasier2022OvercomingCalibration,
    title = {{Overcoming Bias: Equivariant Filter Design for Biased Attitude Estimation With Online Calibration}},
    year = {2022},
    journal = {IEEE Robotics and Automation Letters},
    author = {Fornasier, Alessandro and Ng, Yonhon and Brommer, Christian and Bohm, Christoph and Mahony, Robert and Weiss, Stephan},
    number = {4},
    month = {10},
    pages = {12118--12125},
    volume = {7},
    publisher = {Institute of Electrical and Electronics Engineers Inc.},
    doi = {10.1109/LRA.2022.3210867},
    issn = {23773766},
    arxivId = {2209.12038}
}

@inproceedings{Scheiber2023RevisitingApproach,
    title = {{Revisiting Multi-GNSS Navigation for UAVs – An Equivariant Filtering Approach}},
    year = {2023},
    booktitle = {2023 21st International Conference on Advanced Robotics (ICAR)},
    author = {Scheiber, Martin and Fornasier, Alessandro and Brommer, Christian and Weiss, Stephan},
    month = {12},
    pages = {134--141},
    publisher = {IEEE},
    url = {https://ieeexplore.ieee.org/document/10406552/},
    isbn = {979-8-3503-4229-1},
    doi = {10.1109/ICAR58858.2023.10406552}
}


%% file: bibliography/extra.bib
@IEEEtranBSTCTL{BSTcontrol,
    CTLuse_url = "no",
}

@INPROCEEDINGS{DLmatching,
  author={Michalczyk, Jan and Weiss, Stephan and Steinbrener, Jan},
  booktitle={2025 IEEE/RSJ International Conference on Intelligent Robots and Systems (IROS)}, 
  title={Learning Point Correspondences In Radar 3D Point Clouds For Radar-Inertial Odometry}, 
  year={2025},
  volume={},
  number={},
  pages={3055-3062},
  doi={10.1109/IROS60139.2025.11246451}}

@inproceedings{doer2021radar,
  author={Doer, Christopher and Trommer, Gert F.},
  booktitle={2021 28th Saint Petersburg International Conference on Integrated Navigation Systems (ICINS)}, 
  title={Yaw aided Radar Inertial Odometry using Manhattan World Assumptions}, 
  year={2021},
  volume={},
  number={},
  pages={1-9},
  doi={10.23919/ICINS43216.2021.9470842}
}

@ARTICLE{kim2025,
  author={Kim, Changseung and Bae, Geunsik and Shin, Woojae and Wang, Sen and Oh, Hyondong},
  journal={IEEE Robotics and Automation Letters}, 
  title={EKF-Based Radar-Inertial Odometry With Online Temporal Calibration}, 
  year={2025},
  volume={10},
  number={7},
  pages={7230-7237},
  doi={10.1109/LRA.2025.3575320}}

@INPROCEEDINGS{lessismore,
  author={Huang, Qiucan and Liang, Yuchen and Qiao, Zhijian and Shen, Shaojie and Yin, Huan},
  booktitle={2024 IEEE International Conference on Robotics and Automation}, 
  title={Less is More: Physical-Enhanced Radar-Inertial Odometry}, 
  year={2024},
  volume={},
  number={},
  pages={15966-15972},
  doi={10.1109/ICRA57147.2024.10611471}}

@article{ardea,
doi={https://doi.org/10.1002/rob.21949},
author = {Lutz, Philipp and Müller, Marcus G. and Maier, Moritz and Stoneman, Samantha and Tomić, Teodor and von Bargen, Ingo and Schuster, Martin J. and Steidle, Florian and Wedler, Armin and Stürzl, Wolfgang and Triebel, Rudolph},
title = {ARDEA—An MAV with skills for future planetary missions},
journal = {Journal of Field Robotics},
volume = {37},
number = {4},
pages = {515-551},
doi = {https://doi.org/10.1002/rob.21949},
url = {https://onlinelibrary.wiley.com/doi/abs/10.1002/rob.21949},
eprint = {https://onlinelibrary.wiley.com/doi/pdf/10.1002/rob.21949},
year = {2020}
}

@INPROCEEDINGS { cynoh-2025-icra,
    AUTHOR = { Chiyun Noh and Wooseong Yang and Minwoo Jung and Sangwoo Jung and Ayoung Kim },
    TITLE = { GaRLIO: Gravity enhanced Radar-LiDAR-Inertial Odometry },
    BOOKTITLE = { Proceedings of the IEEE International Conference on Robotics and Automation (ICRA) },
    YEAR = { 2025 },
    MONTH = { May. },
    ADDRESS = { Atlanta },
}

@INPROCEEDINGS{jan_ssrr,
  author={Michalczyk, Jan and Scheiber, Martin and Jung, Roland and Weiss, Stephan},
  booktitle={2023 IEEE International Symposium on Safety, Security, and Rescue Robotics (SSRR)}, 
  title={Radar-Inertial Odometry for Closed-Loop Control of Resource-Constrained Aerial Platforms}, 
  year={2023},
  volume={},
  number={},
  pages={61-68},
  doi={10.1109/SSRR59696.2023.10499937}}


%% file: bibliography/preintegration.bib
@inproceedings{Mourikis2007ANavigation,
    title = {{A Multi-State Constraint Kalman Filter for Vision-aided Inertial Navigation}},
    year = {2007},
    booktitle = {Proceedings 2007 IEEE International Conference on Robotics and Automation},
    author = {Mourikis, Anastasios I and Roumeliotis, Stergios I},
    pages = {3565--3572},
    url = {https://ieeexplore.ieee.org/document/4209642},
    doi = {10.1109/ROBOT.2007.364024}
}

@article{Delama2025EquivariantApproach,
    title = {{Equivariant IMU Preintegration With Biases: A Galilean Group Approach}},
    year = {2025},
    journal = {IEEE Robotics and Automation Letters},
    author = {Delama, Giulio and Fornasier, Alessandro and Mahony, Robert and Weiss, Stephan},
    number = {1},
    pages = {724--731},
    volume = {10},
    publisher = {Institute of Electrical and Electronics Engineers Inc.},
    doi = {10.1109/LRA.2024.3511424},
    issn = {23773766},
    arxivId = {2411.05548}
}

@article{Fornasier2025EquivariantSystems,
    title = {{Equivariant symmetries for inertial navigation systems}},
    year = {2025},
    journal = {Automatica},
    author = {Fornasier, Alessandro and Ge, Yixiao and van Goor, Pieter and Mahony, Robert and Weiss, Stephan},
    pages = {112495},
    volume = {181},
    url = {https://www.sciencedirect.com/science/article/pii/S0005109825003905},
    doi = {https://doi.org/10.1016/j.automatica.2025.112495},
    issn = {0005-1098}
}


%% file: bibliography/radar.bib
@article{Nissov2024DegradationOdometry,
    title = {{Degradation Resilient LiDAR-Radar-Inertial Odometry}},
    year = {2024},
    journal = {Proceedings - IEEE International Conference on Robotics and Automation},
    author = {Nissov, Morten and Khedekar, Nikhil and Alexis, Kostas},
    pages = {8587--8594},
    publisher = {Institute of Electrical and Electronics Engineers Inc.},
    isbn = {9798350384574},
    doi = {10.1109/ICRA57147.2024.10611444},
    issn = {10504729},
    arxivId = {2403.05332}
}

@article{Xu2025IncorporatingSLAM,
    title = {{Incorporating Point Uncertainty in Radar SLAM}},
    year = {2025},
    journal = {IEEE Robotics and Automation Letters},
    author = {Xu, Yang and Huang, Qiucan and Shen, Shaojie and Yin, Huan},
    number = {3},
    month = {3},
    pages = {2168--2175},
    volume = {10},
    doi = {10.1109/LRA.2025.3527344},
    issn = {23773766},
    arxivId = {2402.16082}
}

@article{He2021KalmanManifolds,
    title = {{Kalman Filters on Differentiable Manifolds}},
    year = {2021},
    journal = {arXiv:2102.03804v3},
    author = {He, Dongjiao and Xu, Wei and Zhang, Fu},
    month = {2},
    url = {https://arxiv.org/abs/2102.03804v3},
    arxivId = {2102.03804}
}

@article{Michalczyk2023Multi-StateLandmarks,
    title = {{Multi-State Tightly-Coupled EKF-Based Radar-Inertial Odometry With Persistent Landmarks}},
    year = {2023},
    journal = {Proceedings - IEEE International Conference on Robotics and Automation},
    author = {Michalczyk, Jan and Jung, Roland and Brommer, Christian and Weiss, Stephan},
    pages = {4011--4017},
    volume = {2023-May},
    publisher = {Institute of Electrical and Electronics Engineers Inc.},
    isbn = {9798350323658},
    doi = {10.1109/ICRA48891.2023.10160482},
    issn = {10504729}
}

@article{Michalczyk2022Radar-InertialManoeuvres,
    title = {{Radar-Inertial State-Estimation for UAV Motion in Highly Agile Manoeuvres}},
    year = {2022},
    journal = {2022 International Conference on Unmanned Aircraft Systems, ICUAS 2022},
    author = {Michalczyk, Jan and Schoffmann, Christian and Fornasier, Alessandro and Steinbrener, Jan and Weiss, Stephan},
    pages = {583--589},
    publisher = {Institute of Electrical and Electronics Engineers Inc.},
    isbn = {9781665405935},
    doi = {10.1109/ICUAS54217.2022.9836130}
}

@article{Roumeliotis2002StochasticMeasurements,
    title = {{Stochastic cloning: A generalized framework for processing relative state measurements}},
    year = {2002},
    journal = {Proceedings - IEEE International Conference on Robotics and Automation},
    author = {Roumeliotis, Stergios I. and Burdick, Joel W.},
    pages = {1788--1795},
    volume = {2},
    doi = {10.1109/ROBOT.2002.1014801},
    issn = {10504729}
}

@article{Michalczyk2022Tightly-CoupledOdometry,
    title = {{Tightly-Coupled EKF-Based Radar-Inertial Odometry}},
    year = {2022},
    journal = {IEEE International Conference on Intelligent Robots and Systems},
    author = {Michalczyk, Jan and Jung, Roland and Weiss, Stephan},
    pages = {12336--12343},
    volume = {2022-October},
    publisher = {Institute of Electrical and Electronics Engineers Inc.},
    isbn = {9781665479271},
    doi = {10.1109/IROS47612.2022.9981396},
    issn = {21530866}
}

@article{Michalczyk2024Tightly-CoupledOdometry,
    title = {{Tightly-Coupled Factor Graph Formulation For Radar-Inertial Odometry}},
    year = {2024},
    journal = {IEEE International Conference on Intelligent Robots and Systems},
    author = {Michalczyk, Jan and Quell, Julius and Steidle, Florian and Muller, Marcus G. and Weiss, Stephan},
    pages = {3364--3370},
    publisher = {Institute of Electrical and Electronics Engineers Inc.},
    isbn = {9798350377705},
    doi = {10.1109/IROS58592.2024.10801945},
    issn = {21530866}
}
